\documentclass{article} 
\usepackage[final]{cpal_2026}
\usepackage[utf8]{inputenc} 
\usepackage[T1]{fontenc}    
\usepackage{newunicodechar}
\newunicodechar{，}{,}
\usepackage{xcolor}
\usepackage{hyperref}       
\hypersetup{                
  colorlinks,
  citecolor={red!70!black},
  linkcolor={green!70!black},
  urlcolor={blue!70!black}
}
\usepackage{algorithm}
\usepackage{algorithm}
\usepackage{algpseudocode}

\usepackage{amsmath,amsfonts,bm}

\def\eqref#1{equation~\ref{#1}}

\def\1{\bm{1}}

\DeclareMathAlphabet{\mathsfit}{\encodingdefault}{\sfdefault}{m}{sl}
\SetMathAlphabet{\mathsfit}{bold}{\encodingdefault}{\sfdefault}{bx}{n}

\usepackage{hyperref}
\usepackage{url}

\usepackage{algorithm}
\usepackage{algorithmicx}
\usepackage{algpseudocode}
\usepackage{amsmath}

\usepackage{booktabs}       
\usepackage{amsfonts}       
\usepackage{nicefrac}       
\usepackage{microtype}      
\usepackage{xcolor}         
\usepackage{ragged2e} 

\usepackage{amsmath}
\usepackage{amssymb}
\usepackage{mathtools}
\usepackage{amsthm}
\usepackage{array}
\usepackage{caption}
\usepackage{algorithm}
\usepackage{graphicx}
\usepackage{booktabs}
\usepackage{xcolor}
\usepackage{multirow}
\usepackage{amsfonts}
\usepackage{xcolor}
\usepackage{colortbl}
\usepackage{makecell}
\usepackage{float}
\usepackage{subcaption} 
\usepackage{graphicx}
\usepackage{lipsum}
\usepackage{wrapfig}

\usepackage{algpseudocode}  
\usepackage{amsmath}
\usepackage{makecell} 
\usepackage{arydshln} 
\definecolor{mycolor}{RGB}{235, 235, 235}

\author{
  Mingluo Su\textsuperscript{1}, ~Huan Wang\textsuperscript{1}\thanks{Corresponding author: wanghuan@westlake.edu.cn} \\
  \textsuperscript{1}Westlake University  
}

\newcommand{\ourmethod}{{\tt ROSE}}

\title{ROSE: Reordered SparseGPT for More Accurate One-Shot Large Language Models Pruning}

\begin{document}

\maketitle

\begin{abstract}
Pruning is widely recognized as an effective method for reducing the parameters of large language models (LLMs), enabling more efficient deployment and inference.
One classic and prominent path of LLM one-shot pruning is to leverage second-order gradients (\textit{i.e.}, Hessian), represented by the pioneering work SparseGPT~\citep{sparsegpt}.
However, the predefined left-to-right pruning order in SparseGPT leads to suboptimal performance when the weights exhibit \textit{columnar} patterns.
This paper studies the effect of pruning order under the SparseGPT framework. The analyses lead us to propose \ourmethod, a reordered SparseGPT method that prioritizes weights with larger potential pruning errors to be pruned earlier. 
\ourmethod~first performs pre-pruning to identify candidate weights for removal, and estimates both column and block pruning loss. 
Subsequently, two-level reordering is performed: columns within each block are reordered in descending order of 
column loss, while blocks are reordered based on block loss. We introduce the relative range of block loss as a metric to identify 
\textit{columnar} layers, enabling adaptive reordering across the 
entire model.
Substantial empirical results on prevalent LLMs (LLaMA2-7B/13B/70B, LLaMA3-8B, Mistral-7B) demonstrate that \ourmethod~surpasses the original SparseGPT and other counterpart pruning methods. Our code is available at \url{https://github.com/mingluo-su/ROSE}.
\end{abstract}

\section{Introduction}
Large language models (LLMs)~\citep{thoppilan2022lamda,achiam2023gpt,deepseek,qwen3} have demonstrated remarkable capabilities in natural language understanding and generation attributed to the massive scale of their architectures and training data~\citep{translation,text-summarization,question-answering,feng2025efficient}.
However, with hundreds of billions of parameters, these models require substantial memory and computational resources, posing significant challenges for deployment on resource-constrained devices~\citep{llm-challenge,retrain-1,bai2025ressvd}.

Model pruning~\citep{OBD,OBS,magnitude,zhu2025obs,feng2024oracle,tuo2025sparsessm} is an effective way to enhance model inference and deployment efficiency by removing less critical weights while maintaining competitive performance. 
Traditional pruning methods typically determine which weights to prune in a single pass based on designed criteria~\citep{global-talor,global-hessian,global-fiter-retraining}, or iteratively select the weights with the smallest pruning error for removal, followed by retraining the remaining weights to recover performance~\citep{itertive-channel,magnitude-itertive,itertive-magnitude-retrain,itertive-Growth}. 
%

Nevertheless, retraining-based approaches become prohibitively expensive and time-consuming for LLMs given the significant computational cost required for full-model optimization. On the contrary, pruning approaches for LLMs focus on post-training pruning (PTP) methods~\citep{slimgpt,wanda,sparsegpt,dsnot}. 
Following the classic paradigms of traditional pruning, some current approaches directly prune weights in a single pass but without further adjusting the remaining parameters. Those methods focus on adopting more comprehensive pruning masks~\citep{wanda,dsnot}. 
Another paradigm adjusts the remaining weights by using closed-form second-order solutions~\citep{OBS,frantar2022obc} instead of retraining, represented by the pioneering work SparseGPT~\citep{sparsegpt}.
SparseGPT implements a layer-wise approximate compensation strategy and enables few-hour unstructured pruning of hundred-billion-parameter models, safely pruning up to 60\% of parameters without fine-tuning, highlighting its value for LLMs pruning.

\begin{figure}[!t]
\centering
\resizebox{\textwidth}{!}{%
\begin{tabular}{@{}c@{\hspace{1mm}}c@{\hspace{1mm}}c@{}}
\includegraphics[width=0.32\linewidth]{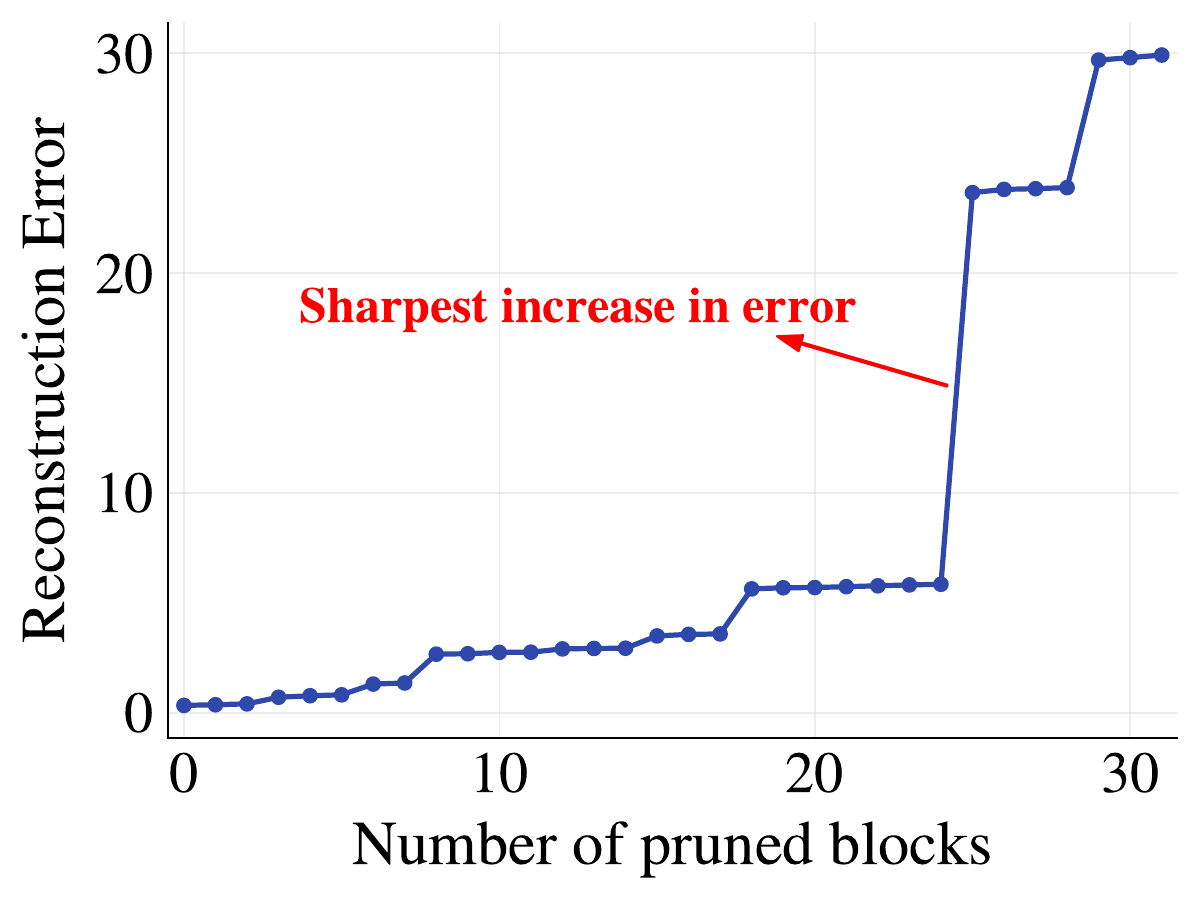}
&
\includegraphics[width=0.32\linewidth,trim=0 0 0 30,clip]{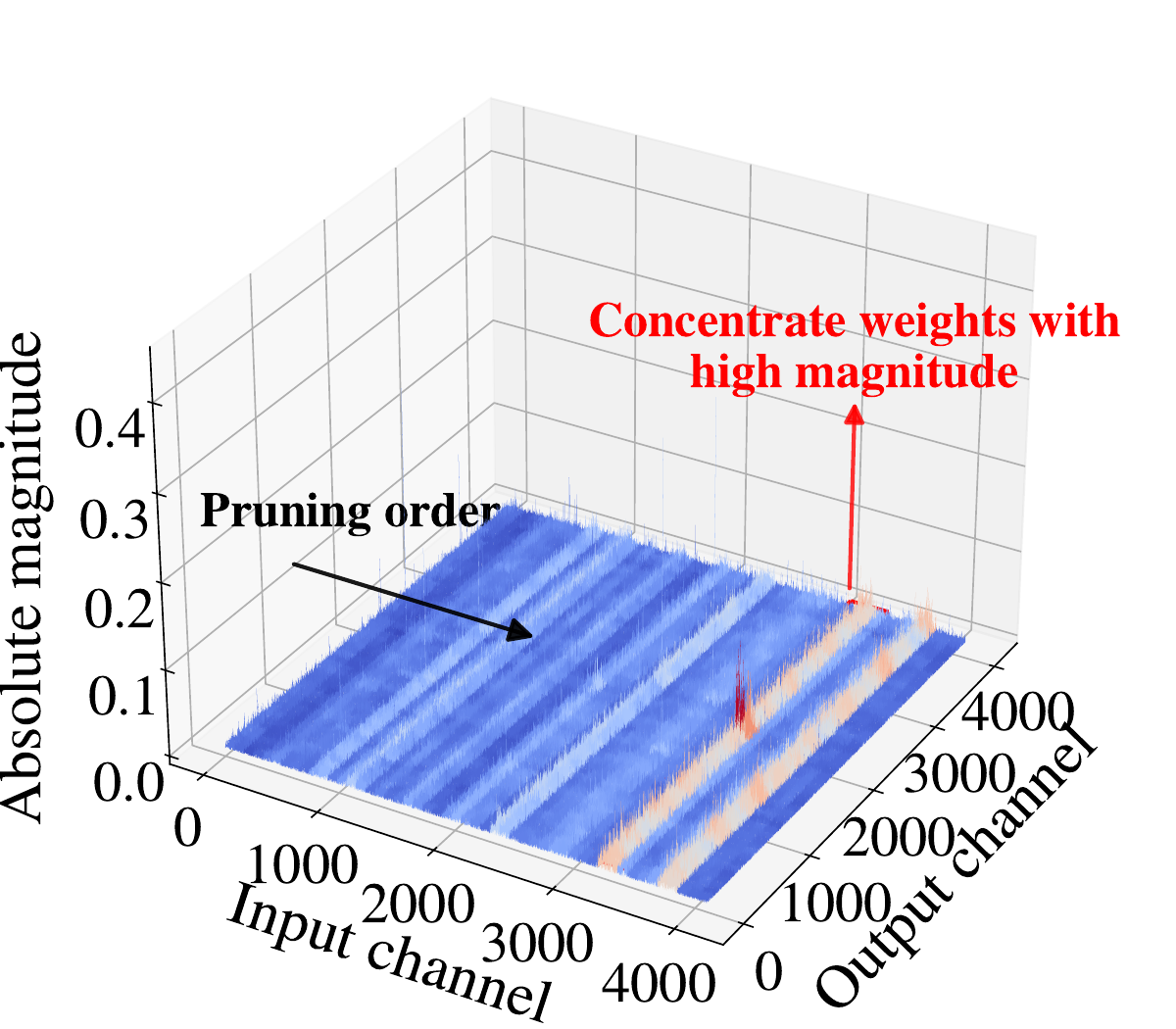}
&
\includegraphics[width=0.32\linewidth]{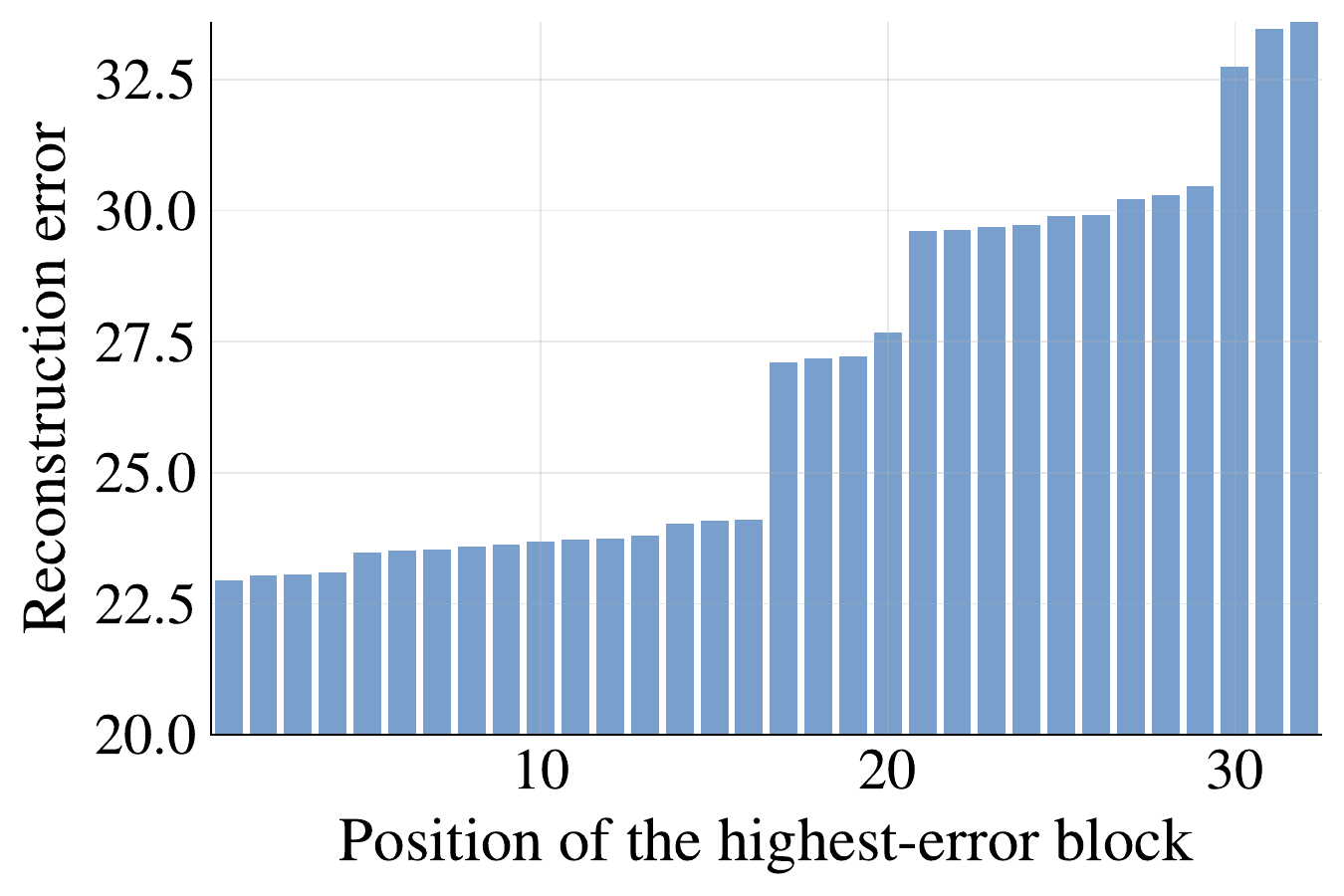}
\\
\vspace{4mm}
(a) \small \makecell[l]{Change in reconstruction error } 
&
(b) \small \makecell[l]{Weight visualization} 
& 
(c) \small \makecell[l]{Reconstruction error after reordering} 
\\
\end{tabular}%
}
\vspace{-3mm}
\caption{\small (a) Change reconstruction error of the "self\_attn.o\_proj" layer in the first Transformer Block of LLaMA2-7B during SparseGPT pruning as the number of pruned blocks increases. The sharpest increase in reconstruction error appears at a later stage.
(b) Weight visualization of the corresponding layer. It exhibits a \textit{columnar} pattern along the input channel, and there is a block with the most concentrated high-magnitude weights as illustrated.
(c) Different reconstruction error after reordering the original block with the highest pruning error. The earlier the original block is pruned, the smaller the reconstruction error.}
\vspace{-3mm}
\label{fig:introduction}
\end{figure}

Revisiting the pruning process in SparseGPT, we find it adopts a fixed block\footnote{Throughout this paper, the lowercase term "block"  refers to a sub-matrix along the input channel of a single layer.} sparsity rate and block size during iterative blocking pruning, and differences in weight distribution across blocks lead to obvious variations in pruning errors. Figure~\ref{fig:introduction}(a) shows the change of layer-wise reconstruction error as the number of pruned blocks increases for the "self\_attn.o\_proj" layer in the first Transformer Block of LLaMA2-7B, where a sharp increase occurs at the late pruning stage, while error increases in other blocks remain relatively moderate. Figure~\ref{fig:introduction}(b) visualizes the weight magnitude of this layer. The weight distribution reveals a \textit{columnar} pattern where weights with similar magnitude are concentrated within blocks. The block with the sharpest increase in pruning error in Figure~\ref{fig:introduction}(a) corresponds precisely to the block with the most concentrated high‑magnitude weights in Figure~\ref{fig:introduction}(b).

Since SparseGPT performs approximate compensation using a common subset of remaining weights, earlier pruned weights have access to more available weights for error correction. The final reconstruction results are influenced by the pruning order. To mitigate this effect, we reorder the block with the highest pruning error from the earliest position to the last, while preserving the relative positions of other blocks. The resulting reconstruction errors under different modified pruning orders are shown in Figure~\ref{fig:introduction}(c).
Interestingly, the earlier this block is pruned, the smaller the final reconstruction error and vice versa.
This observation leads to the following question: \textit{Can we achieve a better weight reconstruction in SparseGPT by proposing an optimized pruning order?}

In this paper, we introduce \ourmethod, a one-shot pruning order adjustment method based on SparseGPT. A pre-pruning step is performed to identify weights that are highly likely to be pruned, based on which both the column-wise losses and block-wise pruning losses are calculated. 
Columns within each block are reordered in descending order of their losses, while blocks are reordered according to their block losses. 
We identify layers exhibiting the \textit{columnar} pattern based on the fluctuation range of block-wise loss and perform reordering for those layers.
The experimental results demonstrate that \ourmethod~can surpass the original SparseGPT and other existing unstructured pruning methods in prevalent LLMs.
Our contributions are as follows:

%
\begin{itemize}
    \item We find that a key factor in accurate one-shot pruning based on the SparseGPT framework is the pruning order, and propose \ourmethod~to study the problem for the first time. 
    
    \item We propose a more optimal pruning order for layers that exhibit a \textit{columnar} pattern and an evaluation metric for detecting layers exhibiting such patterns.
    \item Extensive evaluations on prevalent models suggest our method performs favorably against prior SoTA counterparts.
    
\end{itemize}

\section{Related Work}
\subsection{Network Pruning}
Network pruning aims to reduce redundant parameters while maintaining model accuracy~\citep{OBD,OBS}. 
In terms of workflow, one pruning paradigm is to determine all weights to be pruned based on a certain importance criterion at the initial stage and then retrain the remaining weights to recover performance~\citep{global-fiter-retraining,global-talor,global-hessian}. 
The other is iterative pruning, which involves repeated cycles of pruning based on certain criteria, subsequent fine-tuning of the remaining weights, and re-assessment of the pruning criteria.
Methods in this category generally adopt a greedy order, pruning weights in ascending order of pruning error~\citep{woodfisher,itertive-channel,magnitude-itertive}. 
As pruning progresses, the amount of remaining weight available for compensation gradually decreases. Consequently, in the later stage, when more significant weight is removed, the available parameters in the network become increasingly limited.
To date, no work has explored the impact of pruning order on final model performance.
\subsection{Unstructured Pruning for LLMs}
Unstructured pruning aims to remove unimportant individual weights from the network~\citep{OBD,magnitude}, which differs from structured pruning that aims to remove entire structures such as channels, attention heads, filters, and layers~\citep{head-sixteen,global-fiter-retraining,depgraph,channel-pruning}. 
Unstructured pruning preserves the original model structure and can be performed in a training-free manner~\citep{OBS,frantar2022obc}, an advantage that is particularly critical under conditions of constrained fine-tuning resources in the era of LLMs.
Recent years have witnessed growing attention toward unstructured methods for LLMs. For instance,
SparseGPT \citep{sparsegpt} pioneers one-shot unstructured pruning for LLMs via layer-wise Hessian-based reconstruction for compensation.
Following a different yet simpler paradigm of pruning without weight updates, Wanda \citep{wanda} combines weight magnitude and activation as pruning criteria, while DSnoT~\citep{dsnot} improves upon it by dynamically adjusting the pruning mask.
In contrast, OATS~\citep{oats} approximates each weight matrix as the sum of a sparse matrix and a low-rank matrix, thereby explicitly maintaining the critical outlier features~\citep{outiler-gpt-int8} of LLMs.

\section{Prerequisites}\label{sec:prerequisite}
\textbf{Layer-Wise Pruning.}
Layer-wise pruning~\citep{frantar2022obc,shin2024rethinking}aims to remove less significant weights from each layer sequentially while maintaining overall model performance.
The pruning process for a given layer $l$ is formulated as a minimization problem of the $\ell_2$-error between the original and pruned outputs, defined as follows:
\begin{equation}
\operatorname{argmin}_{\mathbf{\hat{W}}_l}||\mathbf{W}_l \mathbf{X}_l - \mathbf{\hat{W}}_l \mathbf{X}_l||_2^2,
\label{eq:mse}
\end{equation}
where $\mathbf{W}_l$ represents the weight matrix before pruning, $\mathbf{\hat{W}}_l$ is the pruned weight matrix, and $\mathbf{X}_l$ denotes the input to the layer $l$.

\textbf{Optimal Brain Surgeon (OBS) Framework.}
Optimal Brain Surgeon (OBS) is based on a Taylor expansion of the loss function. It removes the weight with minimal impact on the objective function and updates the remaining weights to minimize the change in loss.
Given a weight matrix $\mathbf{W} \in \mathbb{R}^{M \times N}$ and the corresponding input data matrix $\mathbf{X} \in \mathbb{R}^{D \times N}$, let $\mathbf{H}=\mathbf{XX}^{\top}$ denote the corresponding Hessian matrix, the increase in loss $\Delta \mathcal{L} $ caused by the removal of the weight $w_q$ and the optimal updating of the remaining weights $\Delta \mathbf{w} $ are given by:
\begin{equation}
\Delta \mathcal{L} = \frac{w_q^2}{[\mathbf{H}^{-1}]_{qq}}, \quad  
\Delta \mathbf{w} = -\frac{w_q}{[\mathbf{H}^{-1}]_{qq}}\mathbf{H}^{-1}_{:,q}.
\label{eq:obs_error}
\end{equation}
In the process of pruning, OBS employs an iterative update approach that involves multiple Hessian inverse operations, leading to high computational complexity for large-scale models.
To address these problems, OBC \citep{frantar2022obc} decomposes the objective of the layer-wise reconstruction into subproblems by row and proposes an efficient computational framework for matrix inversion through optimized Gaussian elimination. However, its direct application to LLMs remains computationally expensive.

\textbf{Revisiting SparseGPT.}
SparseGPT \citep{sparsegpt} is a one-shot pruning method that adapts the OBS framework specifically for LLMs.
It decouples pruning into mask selection and approximate sparse weight reconstruction. 
For mask selection, SparseGPT selects the pruning mask for $B_S$ columns at a time and adaptively chooses the mask during the pruning process.
For weight reconstruction, it employs a fixed left-to-right pruning order and leverages the common set of remaining pruned weights for multi-row parallel compensation to achieve pruning acceleration. The inverse Hessian information used for weight compensation during the whole pruning process can be stored in the lower triangular matrix $\mathbf{L}$ obtained from the Cholesky decomposition~\citep{frantar2022gptq} of $\mathbf{H}^{-1}$ in advance:
\begin{equation}
\left[ \mathbf{H}_{i:,i:}\right]^{-1} = \mathbf{L}_{i,i}  \mathbf{L}^{\top}_{i:,i:},
\end{equation}
\label{eq:cholesky_h-1}
where $\mathbf{H}^{-1}=\mathbf{LL}^{\top}$.
\section{Methodology}
In this section, we first analyze the dilemma of adjusting the pruning order within the SparseGPT framework and give our main solutions.
Then, we propose \ourmethod: (1) By performing a pre-pruning step, candidate weights with a high probability of being pruned are selected out to estimate both column loss and block. (2) By performing two-level reordering, weights with greater potential errors are pruned earlier. (3)By calculating the relative range of block loss to identify \textit{columnar} layers, an automatic reordering strategy is implemented for whole models.


\subsection{Analyses}
\begin{wrapfigure}{r}{0.32\textwidth} 
    \includegraphics[trim=1mm 5mm 2mm 1mm,clip,width=0.95\linewidth]{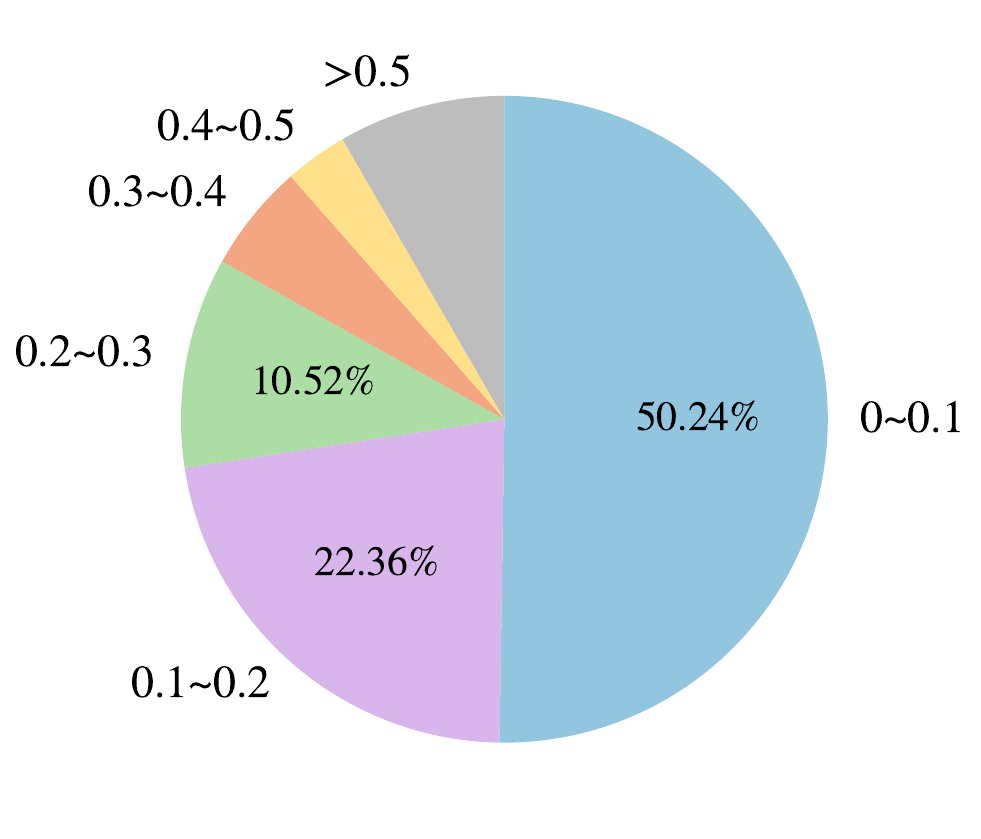}
    \caption{\small The distribution of relative change of weights before and after pruning. The majority of weights remain relatively stable.}
    \label{fig:analysis-delta-w}
\end{wrapfigure}
Our objective is to prioritize weights with larger pruning errors to be pruned earlier. To achieve this, we need to determine which weights will be removed first.
In SparseGPT, once a block is pruned, the mask for the subsequent block is determined based on the updated weights, making it difficult to precisely predict which weights will be pruned.

Fortunately, we observe that most weights (>80\%) deviate only marginally (<30\%) from their original values, as shown in Figure~\ref{fig:analysis-delta-w}. This suggests that the relative importance of most weights remains largely stable throughout the pruning process. Therefore, we can first estimate which weights are highly likely to be pruned based on the magnitudes of the initial weights, compute the corresponding pruning loss, and then reorder the weights accordingly.
Additionally, since SparseGPT adopts a block-wise masking strategy, we reorder each block as a whole to ensure that the mask remains essentially unchanged.

\subsection{Proposed \ourmethod}

\begin{figure}[t]
\centering
  \includegraphics[trim=33mm 20mm 23mm 0mm,clip,width=\linewidth]{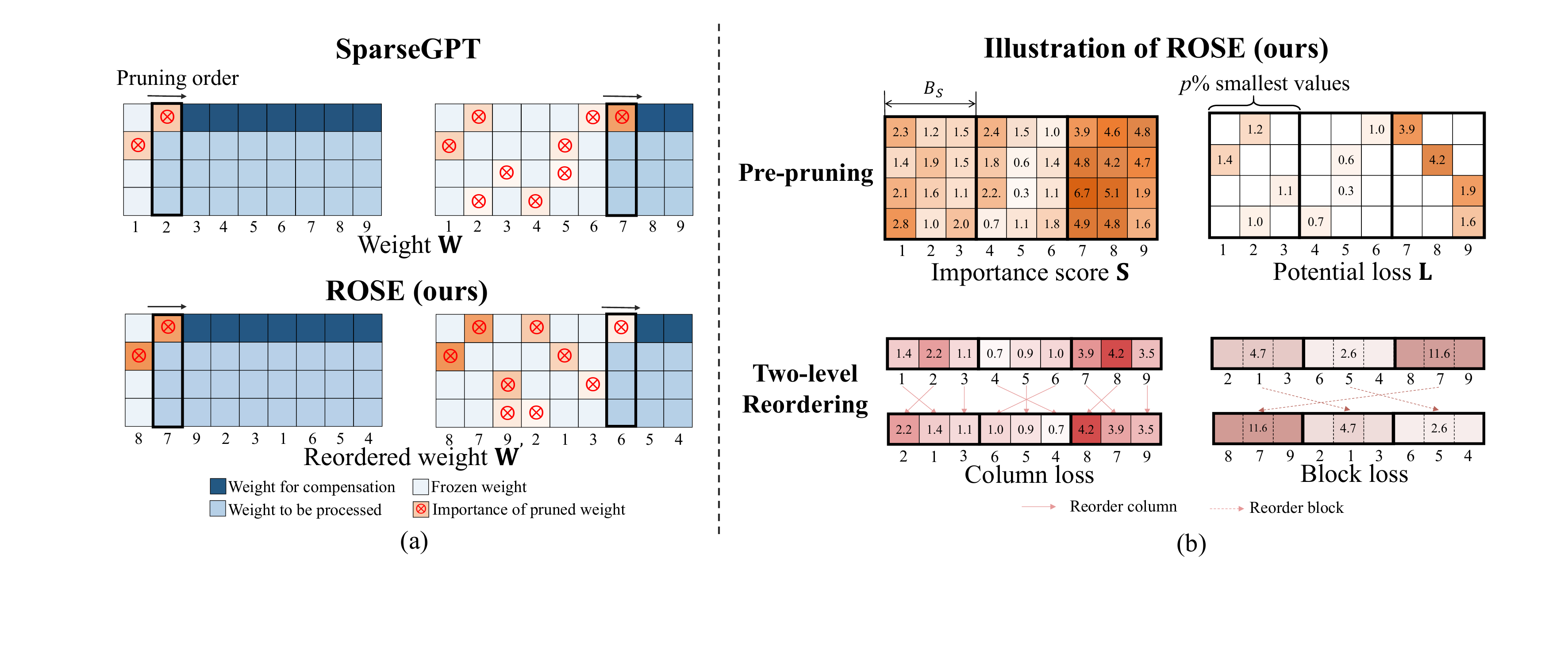}
\vspace{-5mm}
\label{fig:overview}
\definecolor{myorange}{RGB}{216, 98, 19}   
\definecolor{blue}{RGB}{36,84,126}   

\caption{\small(a) Overview of difference between SparseGPT and \ourmethod. \textcolor{myorange}{Orange} color represents weight importance, and the darker the color, the greater the importance. In SparseGPT, the number of weights available for error compensation (shown in \textcolor{blue}{dark blue}) decreases during pruning, limiting recovery if high-error weights are pruned late.
\ourmethod~reorders those with potentially large pruning errors to the front to be pruned earlier. In this way, more parameters remain available for larger error compensation. (b) Illustration of our \ourmethod~for the \textit{columnar} layer.
Given the dense weight $\mathbf{W}$ and target sparsity rate $p\%$, we calculate the importance score $\mathbf{S}$ and split it into blocks based on $B_S$. 
The smallest $p\%$ of values from each block are selected as the loss matrix $\mathbf{L}$. Column loss and block loss are calculated based on the loss matrix.
Columns within one block are reordered in descending order of column loss, and blocks are reordered in descending order of block loss.
}
\end{figure}

\textbf{Pre-pruning.} We estimate the potential pruning loss by performing a pre-pruning step based on the importance of the initial weights.
For the importance score for each weight, we adopt the metric proposed in Wanda~\citep{wanda}, which combines the weight magnitude and the corresponding input activation.
Specifically, given the weight matrix $\mathbf{W} \in \mathbb{R}^{M \times N}$, where $M$ donates the number of output channels and $N$ donates the number of input channels,
let $\mathbf{X} \in \mathbb{R}^{(B \times L) \times N}$ represent the input activation matrix, where $B$ represents the batch size and $L$ denotes the sequence length.
For a single weight $\mathbf{W}_{ij}$, the importance score $\mathbf{S}_{ij}$ is defined as:
\begin{equation}
\mathbf{S}_{ij} = |\mathbf{W}_{ij}| \cdot \|\mathbf{X}_j\|_2, 
\label{eq:weight score}
\end{equation}
where $\cdot$ represents the element-wise product and $\|\mathbf{X}_j\|_2$ denotes the $\ell_2$ norm of the corresponding input activation. 

We construct the potential loss matrix~$\mathbf{L}$ by performing the pre-pruning process.
Following the fixed iterative block pruning of SparseGPT, $\mathbf{W}$ is divided into $K$ blocks where $K = \lceil N / B_s \rceil$ along the column dimension. The corresponding weight blocks and input activation blocks can be denoted as $\mathbf{W}^{(k)} = \mathbf{W}[:,i_1:i_2]$ and $\mathbf{X}^{(k)} = \mathbf{X}[:,i_1:i_2]$, respectively, 
where $i_1 = (k-1) \cdot B_s$ and $i_2 = \min(k \cdot B_s, N)$.
Then we can get block score $\mathbf{S}^{(k)}$ by Equation~\ref{eq:weight score}.
Let $p\%$ be the target sparsity rate. For each block $k = 1, 2, \ldots, K$, the smallest $p\%$ of elements in $\mathbf{S}^{(k)}$ are extracted to form candidate pruning loss of blocks~$\mathbf{L}^{(k)}$. 

\textbf{Two-level Reordering.}
In order to prioritize pruning weights with larger errors, our reordering process includes two levels: column reordering and block reordering. Column reordering is performed within each block separately. Specifically, the pruning loss of each column in block $k$ is calculated as $l_{j}^{(k)}=\sum_{i=1}^{M} \left[ \mathbf{L}^{(k)} \right]_{ij}$.
The columns in each $\mathbf{W}^{(k)}$ are sorted in descending order of column loss $l_{j}^{(k)}$:
\begin{equation}
\mathbf{W}^{(k)}\gets \left[\mathbf{w}_{j_1}^{(k)},\mathbf{w}_{j_2}^{(k)},\dots,\mathbf{w}_{j_B}^{(k)}\right] \quad \text{where} \quad l_{j_1}^{(k)} \ge l_{j_2}^{(k)} \ge \dots \ge l_{j_B}^{(k)}.
\label{eq:reorder-column}
\end{equation}
 For block reordering, the entire block is treated as a unit and reordered in descending order.The total block loss is calculated as $L^{(k)} = \sum_{i=1}^{M} \sum_{j=i_1}^{i_2} \left[ \mathbf{C}^{(k)} \right]_{ij}.$
All blocks are reordered in descending order based on $L^{(k)}$:
\begin{equation}
\mathbf{W}\gets\left[\mathbf{W}^{(k_1)},\mathbf{W}^{(k_2)}, \dots, \mathbf{W}^{(k_K)}\right]  \quad \text{where} \quad L^{(k_1)} \ge L^{(k_2)} \ge \dots \ge L^{(k_K)}.
\label{eq:reorder-block}
\end{equation}
Through these two-levels reordering operations, weights with larger pruning errors are prioritized to be pruned earlier.

\textbf{Columnar Layer Identification.}
Reordering is performed on those layers that presented the \textit{columnar} pattern. In order to identify this structure, we use the differences in block losses for identification. To quantify this difference, we define the relative range of block loss as:
\begin{equation}
R_{\text{rel}} = \frac{\max_k {L}^{(k)} - \min_k {L}^{(k)}}{\text{mean}~{{L}^{(k)}}}.
\label{eq:rr}
\end{equation}
Consequently, if the relative range of block loss of a layer exceeds a predefined threshold, we classify it as \textit{columnar} layer and apply reordering to it.

%
\clearpage
\section{Experimental Results}
\subsection{Experiment Settings}
\textbf{Models and Datasets}.
We select the current public LLMs for evaluation, including the LLaMA2 series~\citep{touvron2023bllama}, LLaMA3 series~\citep{llama3}, and Misrtal-7B~\citep{mistral}.
%
These models range in size from 7 billion to 70 billion parameters.
We primarily assess the performance of pruned large language models through perplexity, a widely adopted and stable metric for measuring LLM performance， and use WikiText-2-raw \citep{merity2016pointer} datasets. 
To further evaluate the capabilities of pruned models, we also conduct experiments on seven standard common-sense benchmark tasks: BoolQ \citep{clark2019boolq}, WinoGrande \citep{sakaguchi2021winogrande}, PIQA~\citep{piqa}, OpenBookQA \citep{obqa}, HellaSwag~\citep{zellers2019hellaswag}, ARC-Easy and ARC-Challenge \citep{arc}.
All zero-shot tasks are performed uniformly based on the lm-eval-harness framework~\citep{lm-eval}.

\textbf{Comparison Methods.} We compare our method with counterpart methods, including: (1) Magnitude pruning~\citep{magnitude} that removes weights based on the magnitude metric. (2) SparseGPT \citep{sparsegpt} that utilizes approximate second-order Hessian information to evaluate weight importance and perform weight reconstruction. (3) Wanda \citep{wanda} that removes weights based on magnitudes multiplied by the corresponding input activation norms. (4) DSnoT \citep{dsnot} that performs training-free fine-tuning with dynamic masks after pruning. (5) OATS \citep{oats} that decomposes weight matrices into a sparse matrix and a low-rank matrix, with a designed strategy to preserve critical outlier features.

\textbf{Implementation Details.}  
\ourmethod~is implemented with the PyTorch framework~\citep{pytorch} and HuggingFace Transformers \citep{huggingface}. Consistent with previous works \citep{sparsegpt}, the calibration data contains 128 samples randomly selected from the first shard of the C4 dataset~\citep{c4}. Each sample contains sequences of 2048 tokens.
All experiments are conducted on NVIDIA 48GB 4090 GPUs. 
The results of comparison methods are reproduced by their official code. DSnoT is combined with both Magnitude, Wanda, and SparseGPT in its paper. In our experiments, we run them both and take the best one to put in the results. For SparseGPT and~\ourmethod, the blocksize is set to 128. For \ourmethod, the threshold for the \textit{columnar} layer is set to 0.5, and the specific reasons are explained in Appendix~\ref{sec:Weigh Visualization}.
\subsection{Reconstruction Error Analyses}

In this section, we evaluate the reconstruction error of the single layer using different methods.
First, we analyze the reconstruction error by all methods. The results are shown in Table~\ref{tab:error-all-methods}.
It can be observed that Wanda, DSnoT, and Magnitude, which prune weights directly, exhibit a significant increase in reconstruction error as sparsity increases. OATS, despite decomposing the weights into the sum of a sparse matrix and a low-rank matrix, also shows large reconstruction errors. 
In contrast, SparseGPT and \ourmethod~consistently achieve lower reconstruction errors across different sparsity levels. This is because both methods are based on the OBS second-order effective compensation. Moreover, our method achieves a smaller reconstruction error than SparseGPT at all sparsity rates.

\begin{figure}[!t]
\centering
\begin{tabular}{@{}cc@{}}
     \includegraphics[width=0.46\linewidth]{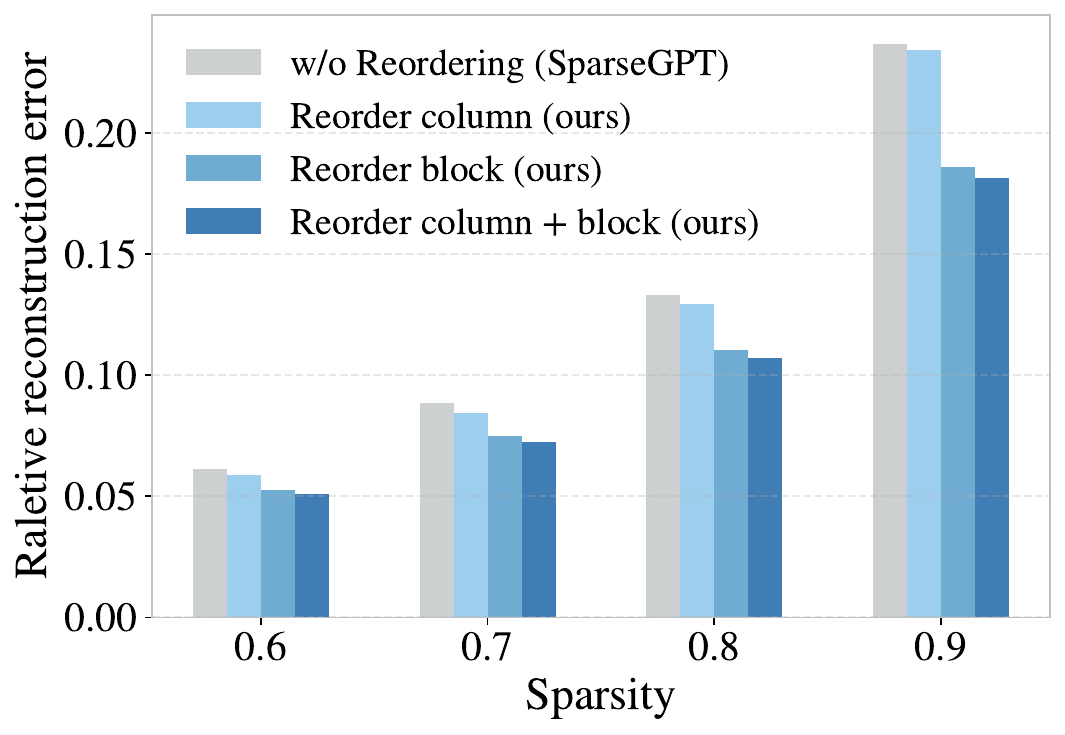} &
     \includegraphics[width=0.46\linewidth]{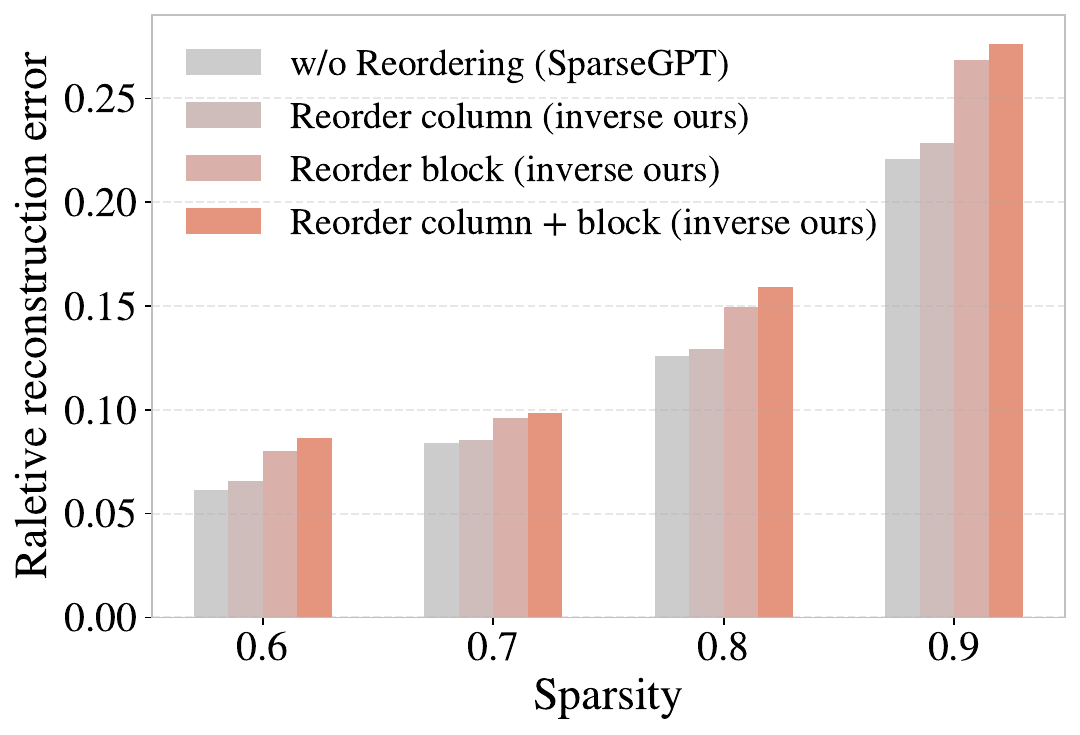} \\
     (a) \small Descending order & (b) \small Ascending order
\end{tabular}
\caption{Relative reconstruction error of the "self\_attn.o\_proj" layer in the second Transformer Block of LLaMA2-7B by \ourmethod~and its variants at varying sparsity rates.}
\label{fig:error-ours}
\end{figure}

Figure~\ref{fig:error-ours} provides a detailed analysis demonstrating that~\ourmethod~achieves a lower reconstruction error than SparseGPT.
As shown in Figure~\ref{fig:error-ours}(a), both column reordering and block reordering individually contribute to reducing the reconstruction error, with block reordering yielding more pronounced improvements.
Meanwhile, the reduction in reconstruction error becomes more pronounced as the sparsity level increases.
Interestingly, Figure~\ref{fig:error-ours}(b) shows the opposite trend when the order of our method is reversed, pruning the blocks and columns first with smaller errors. 
This contrast strongly indicates that the pruning order accounts for pruning error.


\begin{table}[!h]
\centering
\caption{Relative reconstruction error of the "self\_attn.o\_proj" layer in the second Transformer Block of LLaMA2-7B by different methods.}
\scriptsize
\vspace{2mm}
\setlength{\tabcolsep}{7pt}
\resizebox{0.96\textwidth}{!}{%
\begin{tabular}{c c c c c c c }
\toprule
\textbf{Sparsity} & 
\textbf{Magnitude} & 
\textbf{SparseGPT} & 
\textbf{Wanda} & 
\textbf{DSnoT} &
\textbf{OATS} &
\textbf{\ourmethod~(ours)} \\
\midrule

60\% & 1.50e-1 & 6.12e-2 & 9.39e-2 & 9.39e-1 & 1.50e-1 & \textbf{5.09e-2} \\
70\% & 2.20e-1 & 8.84e-2 & 1.40e-1 & 1.40e-1 & 2.00e-1 & \textbf{7.22e-2} \\
80\% & 3.20e-1 & 1.33e-1 & 2.20e-1 & 2.20e-1 & 3.00e-1 & \textbf{1.07e-1} \\
90\% & 5.00e-1 & 2.37e-1 & 3.60e-1 & 3.60e-1 & 4.80e-1 & \textbf{1.81e-1} \\

\bottomrule  
\end{tabular}}
\label{tab:error-all-methods}
\end{table}

\vspace{1mm}
\subsection{Main Benchmark Results}\label{sec:benchmark}
\textbf{Unstructured Pruning Results.}
Table~\ref{tab: ppl-sparsity} shows the WikiText perplexity performance on LLaMA3-8B and Mistral-7B at different sparsity rates.
Under high sparsity conditions, SparseGPT and \ourmethod~clearly outperform other approaches since they adjust the unpruned weights for error compensation. Moreover, \ourmethod~ achieves lower perplexity compared with SparseGPT at most sparsity rates. For example, \ourmethod~reduces perplexity from 203.45 to 172.14 at 80\% sparsity rate on LLaMA3-8B. 

Table~\ref{tab:llama2-0.7} presents the WikiText perplexity performance and zero-shot task performance of unstructured pruning methods on LLaMA2 models at 70\% sparsity rate. SparseGPT and \ourmethod~outperform other pruning methods in most cases. Moreover, \ourmethod~achieves lower perplexity results than the SparseGPT across all evaluated models. 
In terms of zero-shot evaluations, \ourmethod~achieves better average accuracy across all evaluated models than SparseGPT.
For task-specific evaluations, \ourmethod~can achieve higher accuracy in the majority of tasks across the models of different sizes. Notably, at the 7B, our approach surpasses SparseGPT by over 1.5\% in ARC-c and  ARC-e tasks.

\begin{table}[!t]
\centering
\caption{WikiText perplexity (\textcolor{red}{$\downarrow$}) performance on LLaMA3-8B and Mistral-7B model at varying sparsity rates.}
\vspace{2mm}
\setlength{\tabcolsep}{6pt}
\resizebox{0.96\textwidth}{!}{
\begin{tabular}{l *{8}{>{\centering\arraybackslash}p{1.2cm}}}
\toprule
\multirow{2}{*}{\textbf{Method}} & \multicolumn{4}{c}{\textbf{LLaMA-3 8B} (Dense: 6.14)} & \multicolumn{4}{c}{\textbf{Mistral-7B} (Dense: 5.32)} \\
\cmidrule(lr){2-5} \cmidrule(lr){6-9}
 & \textbf{60\%} & \textbf{70\%} & \textbf{80\%} & \textbf{90\%} & \textbf{60\%} & \textbf{70\%} & \textbf{80\%} & \textbf{90\%} \\

\midrule
Magnitude &3.38e5 & 1.62e6 &8.54e6 & 2.35e6 & 31.42 & 8.88e3 & 1.34e4 &1.22e5  \\
SparseGPT  & \textbf{15.23} &\underline{40.48} & \underline{203.45} &\underline{1.10e3} & \underline{9.37} &\underline{21.48} &\textbf{78.69} & \underline{286.54}  \\
Wanda &23.34 & 123.78 & 986.97 & 1.02e4 & 11.11 &57.31 & 236.17 & 5.16e3\\
DSnoT    & 19.66 & 126.99 & 995.57& 8.38e4 & 9.67 & 30.51 &1.91e3&7.76e3 \\
OATS    & 16.34 & 88.93 & 770.54 & 7.95e3 &10.54 &35.20 & 261.60& 6.44e3  \\
\rowcolor{mycolor} 
\textbf{\ourmethod}~(ours) & \underline{15.50} & \textbf{40.29} & \textbf{172.14} & \textbf{840.10} & \textbf{9.30} & \textbf{20.86} &\underline{78.96} & \textbf{266.88} \\
\bottomrule
\end{tabular}}

\label{tab: ppl-sparsity}
\end{table}


\textbf{Semi-structured Pruning Results.}
\ourmethod~can be extended to semi-structured pruning by changing the pre-pruning step according to the semi-structured sparsity pattern and adjusting the blocksize parameter accordingly. 
Specifically, the blocksize is set to 4 for the 2:4 pattern and to 8 for the 4:8 pattern.
Table~\ref{tab:llama-2:4} and~\ref{tab:llama-4:8} show the perplexity performance of LLaMA models in two semi-structured patterns. 
The results show that our approach outperforms SparseGPT in both the 2:4 and 4:8 patterns. 
For example, under the 2:4 pattern on the LLaMA3-8B, our method reduces perplexity on WikiText by 0.5 compared to SparseGPT, demonstrating the effectiveness and superiority of \ourmethod~in semi-structured pruning.
\clearpage

\begin{table}[!t]

\centering
\scriptsize
\caption{WikiText perplexity (\textcolor{red}{$\downarrow$}) and zero-shot task accuracy~(\textcolor{red}{$\uparrow$}) performance on LLaMA2 models at 70\% sparsity rate for different unstructured pruning methods. 
}
\vspace{2mm}
\label{tab:pruning_results}
\setlength{\tabcolsep}{1pt}
\resizebox{0.98\textwidth}{!}{
\begin{tabular}{c c c c c c c c c c c}
\toprule
\multirow{2}{*}{\textbf{Model}} & 
\multirow{2}{*}{\textbf{Method}} & 
\multicolumn{1}{c}{\multirow{2}{*}{\textbf{Perplexity}}} & 
\multicolumn{8}{c}{\textbf{Zero-shot Accuracy}} \\
\cmidrule(lr){4-11}  
 & &  & 
\textbf{BoolQ} & 
\textbf{WinoG.} & 
\textbf{PIQA} & 
\textbf{OBQA} & 
\textbf{HellaS.} & 
\textbf{ARC-e} & 
\textbf{ARC-c} & 
\textbf{Avg.} \\

\midrule
\multirow{7}{*}{LLaMA2-7B}
& Dense & 5.47 & 77.68 & 69.14 & 79.05 & 44.20 & 76.01 & 74.54 & 46.33 & 66.71 \\
\cmidrule(l){2-11} 
& Magnitude & 4.98e4 & 37.95 & 49.25 & 51.52 & 28.00 & 26.32 & 27.90 & \textbf{26.96} & 35.41 \\
& SparseGPT &\underline{27.68} & \underline{63.61} & \underline{58.41} & {62.35} & {29.60} & \underline{40.38} & 40.19 & 23.46 & 45.43 \\
& Wanda & 72.58 & 48.50 & 49.33 & 53.86 & 25.80 & 30.21 & 30.64 & 21.33 & 37.10 \\
& DSnoT & 60.44 & {62.14} & 55.25 & \textbf{63.00} & \underline{30.40} & 39.24 & \textbf{44.15} & \underline{25.94} & \underline{45.73} \\
& OATS & 50.44 & 60.46 & 51.38 & 55.11 & 28.20 & 32.32 & 32.45 & 21.59 & 40.22  \\
& \cellcolor{mycolor}\ourmethod~(ours) & \cellcolor{mycolor}\textbf{26.38} & \cellcolor{mycolor}\textbf{64.04} & \cellcolor{mycolor}\textbf{59.19} & \cellcolor{mycolor}\underline{62.84} & \cellcolor{mycolor}\textbf{30.60} & \cellcolor{mycolor}\textbf{41.35} & \cellcolor{mycolor}\underline{41.71} & \cellcolor{mycolor}25.26 & \cellcolor{mycolor}\textbf{46.43} \\

\midrule
\multirow{7}{*}{LLaMA2-13B}
& Dense  &4.88 & 80.55 & 71.98 & 80.52 & 45.20 & 79.36 & 77.53 & 48.98 & 69.16 \\
\cmidrule(l){2-11} 
& Magnitude & 2.14e2& 38.65 & 49.49 & 53.10 & 26.60 & 29.51 & 32.11 & 24.49 & 36.28 \\
& SparseGPT & \underline{19.78}& \textbf{68.17} & \underline{61.88} & \underline{67.85} & 32.20 & 46.70 & 48.11 & \textbf{28.67} & \underline{50.51} \\
& Wanda &46.22 & 62.08 & 50.75 & 57.34 & 28.20 & 31.62 & 35.56 & {21.25} & 40.97 \\
& DSnoT &31.21 & 64.86 & 56.20 & 66.92 & \textbf{33.60} & \underline{47.17} & \underline{49.45} & 27.22 & 49.35 \\
& OATS &40.80 & 62.42 & 56.04 & 60.39 & 29.20 & 35.27 & 38.01 & 22.61 & 43.42 \\
& \cellcolor{mycolor}\ourmethod~(ours)& \cellcolor{mycolor}\textbf{19.54} & \cellcolor{mycolor}\underline{65.90} & \cellcolor{mycolor}\textbf{63.22} & \cellcolor{mycolor}\textbf{68.01} & \cellcolor{mycolor}\underline{33.00} & \cellcolor{mycolor}\textbf{47.61} & \cellcolor{mycolor}\textbf{49.54} & \cellcolor{mycolor}\underline{27.99} & \cellcolor{mycolor}\textbf{50.75} \\
\midrule
\multirow{7}{*}{LLaMA2-70B}
& Dense & 3.32& 83.76 & 77.98 & 82.70 & 48.80 & 83.81 & 81.06 & 57.25 & 73.62  \\
\cmidrule(l){2-11}
& Magnitude &423.46 & 39.57 & 57.14 & 67.63 & 35.80 & 57.20 & 54.55 & 34.56 & 49.49\\
& SparseGPT & {9.34} & \textbf{80.58} & \textbf{75.30} & \underline{77.04} & {41.60} & {69.19} & \underline{70.03} & \textbf{43.86} & \underline{65.37}\\
& Wanda &10.59& 74.10 & 74.03 & 75.63 & 40.00 & 64.73 & 69.99 & 40.78 & 62.75  \\
& DSnoT  & \textbf{8.29}& 79.02 & 73.95 & \textbf{77.31} & \textbf{42.80} & \textbf{71.96} & 69.53 & 42.75 & 65.33  \\
& OATS &{9.97}& 75.60& 73.23& 75.83&40.70& 68.13&69.49& 41.38 &63.48 \\
& \cellcolor{mycolor}\ourmethod~(ours) & \cellcolor{mycolor}\underline {9.29} & \cellcolor{mycolor}\underline{80.18} & \cellcolor{mycolor}\underline{75.14} & \cellcolor{mycolor}76.44 & \cellcolor{mycolor}\underline{42.60} & \cellcolor{mycolor}\underline{69.40} & \cellcolor{mycolor}\textbf{70.79} & \cellcolor{mycolor}\underline{43.77} & \cellcolor{mycolor}\textbf{65.47} \\
\bottomrule
\end{tabular}}
\label{tab:llama2-0.7}
\end{table}


\begin{table}[!t]
\centering
\begin{minipage}{0.46\textwidth}
\caption{WikiText perplexity (\textcolor{red}{$\downarrow$}) on LLaMA models with 2:4 pattern.}
\vspace{1mm}
\label{tab:llama-2:4}
\setlength{\tabcolsep}{7pt}
\resizebox{\textwidth}{!}{
\begin{tabular} 
{cccc}
\toprule
\textbf{Method} & \textbf{2-7B} & \textbf{2-13B}  &\textbf{3-8B}\\
\midrule
SparseGPT & 11.00   & 8.77 & 16.33 \\
\cellcolor{mycolor}\ourmethod~(ours)& \cellcolor{mycolor}\textbf{10.73} & \cellcolor{mycolor}\textbf{8.60} & \cellcolor{mycolor}\textbf{15.84} \\
\bottomrule
\end{tabular}
}
\end{minipage}
\hspace{6mm}
\begin{minipage}{0.46\textwidth}
\centering
\caption{WikiText perplexity (\textcolor{red}{$\downarrow$}) on LLaMA models with 4:8 pattern.}
\vspace{1mm}
\label{tab:llama-4:8}
\setlength{\tabcolsep}{7pt}

\resizebox{\textwidth}{!}{
\begin{tabular}{cccc}
\toprule
\textbf{Method} & \textbf{2-7B} & \textbf{2-13B}  &\textbf{3-8B}\\
\midrule
SparseGPT & 8.46  & 7.00  & 12.20\\
\cellcolor{mycolor}\ourmethod~(ours) & \cellcolor{mycolor}\textbf{8.30} & \cellcolor{mycolor}\textbf{6.96} & \cellcolor{mycolor}\textbf{12.00} \\
\bottomrule
\end{tabular}

}
\end{minipage}
\vspace{-2mm}
\end{table}

\subsection{Ablation Study}

\begin{figure}[!t]
\centering
\begin{tabular}{@{}ccc@{}}
     \includegraphics[width=0.3\linewidth]{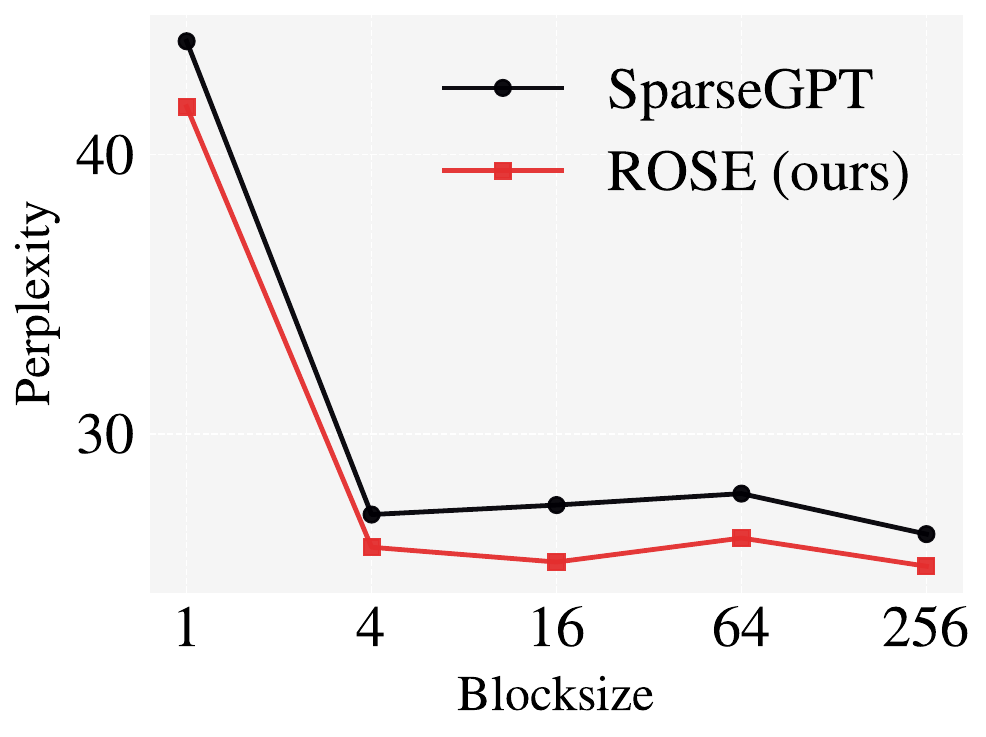} &
     \includegraphics[width=0.3\linewidth]{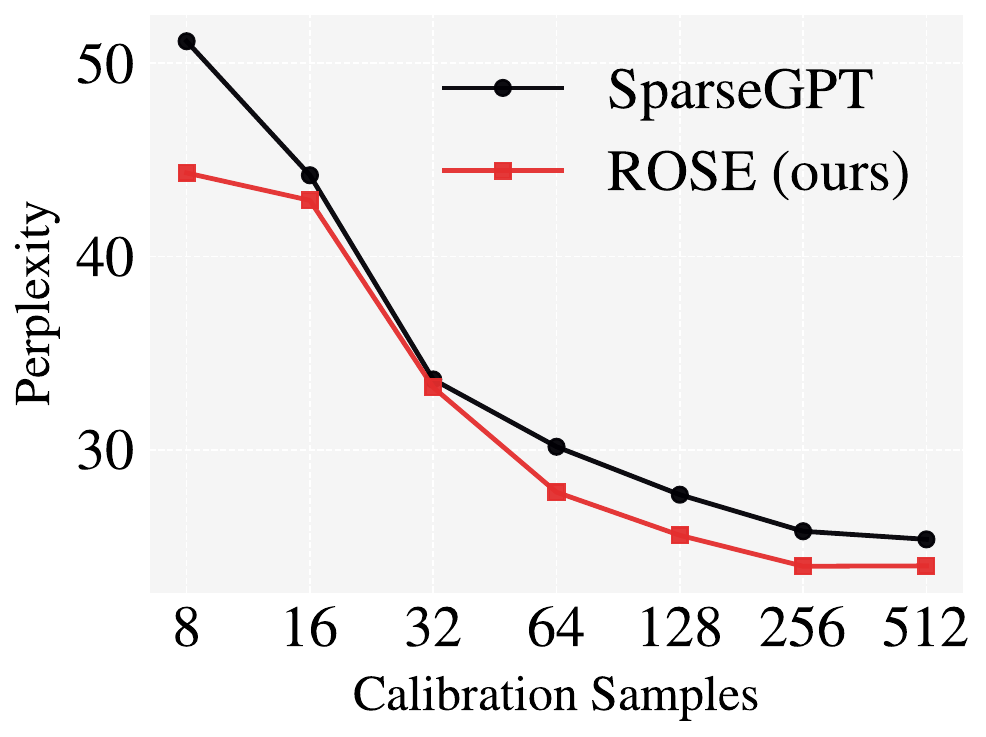} & 
     \includegraphics[width=0.3\linewidth]{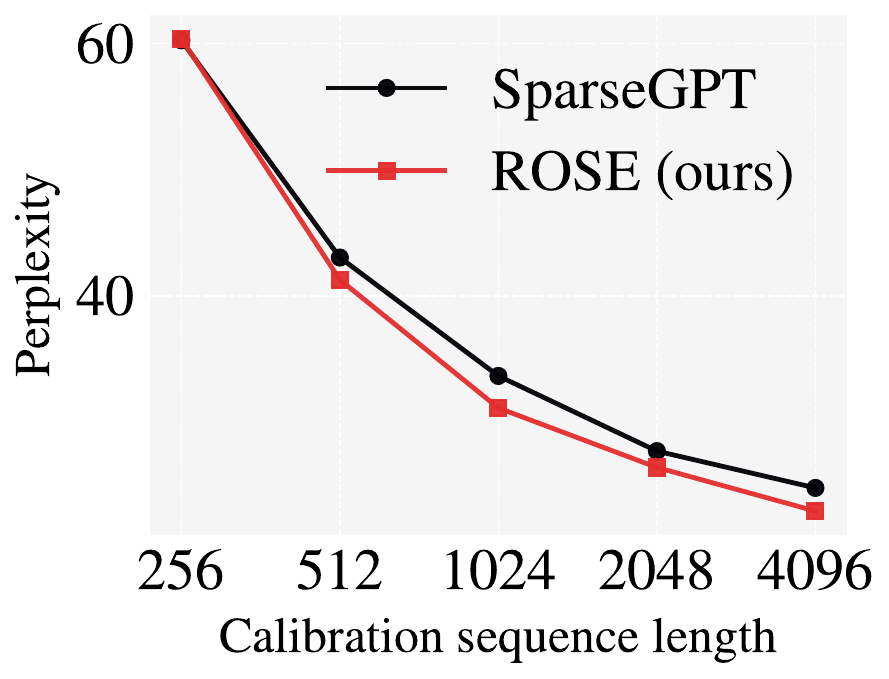} 
     \end{tabular}
\caption{Ablation study of blocksize, calibration samples, and calibration sequence length in LLaMA2-7B at 70\% sparsity rate.}
\label{fig:ablation}
\end{figure}

\textbf{Blocksize.} \ourmethod~involves reordering both blocks and the columns within one block, a process governed by the blocksize. 
This section investigates its performance with SparseGPT under identical blocksize conditions on LLaMA2-7B at 70\% sparsity rate. 
As illustrated in the Figure~\ref{fig:ablation}(a), \ourmethod~exhibits robustness similar to SparseGPT, with WikiText perplexity remaining stable over a wide range of blocksize values and \ourmethod~consistently achieves a lower perplexity than SparseGPT.

\textbf{Calibration Data.}
Since both SparseGPT and \ourmethod~rely on the Hessian matrix computed from calibration data for weight compensation, we analyze the impact of calibration data number and sequence length on the LLaMA2-7B model at 70\% sparsity rate.
The results are shown in Figure~\ref{fig:ablation}(b) and~\ref{fig:ablation}(c).
For the number of calibration data, \ourmethod~consistently outperforms SparseGPT by achieving lower perplexity. 
Similarly, longer input sequences also lead to reduced perplexity, and \ourmethod~maintains its performance advantage across all sequence lengths.

\vspace{-6mm}
\begin{table}[!t]
\centering
\footnotesize 
\setlength{\tabcolsep}{4pt}  
\caption{Pruning time (minutes) for pruning a whole model with a calibration set of size 128 at 70\% sparsity rate. }
\resizebox{0.78\textwidth}{!}{
\begin{tabular}{ccccccc}
\toprule
\textbf{Model} &\textbf{Magnitude} & \textbf{SparseGPT} & \textbf{Wanda} & \textbf{DSnoT} & \textbf{OATS} & \textbf{\ourmethod~(ours)} \\ 
\midrule
LLaMA2-7B &-& 4.76 & 1.75 & 1.95 & 572& 5.15 \\
LLaMA2-13B&-& 8.45& 1.85 & 2.35 & 925& 9.34\\
\bottomrule
\end{tabular}}
\label{tab:time and gpus}
\end{table}

\subsection{Running Consumption Analyses}

\textbf{Pruning Time Consumption.} Table \ref {tab:time and gpus} compares the pruning computations of \ourmethod~and other counterpart pruning methods on LLaMA2-7B and 13B models. Wanda and DSnoT do not involve any weight updates, leading to extremely fast pruning speeds. However, they suffer from significant performance degradation at high sparsity rates, as demonstrated in Section~\ref{sec:benchmark}. OATS exhibits the longest pruning time. For instance, it requires 572 minutes to prune the LLaMA2-7B model, which is hundreds of times that of \ourmethod. \ourmethod~introduces two additional lightweight steps compared to SparseGPT: computing the pruning loss and performing reordering operation. The overall pruning time increases marginally relative to SparseGPT (from 4.76 minutes to 5.15 minutes on the LLaMA2-7B and from 8.45 minutes to 9.34 minutes on the LLaMA2-13B, respectively). 
\begin{wraptable}{r}{0.34\textwidth}
    \centering
    \vspace{3mm}
       \caption{\small End-to-end latency obtained by different methods on LLaMA2-70B.}
       \vspace{-1mm}
    \resizebox{0.34\textwidth}{!}{
    \begin{tabular}{ccc}
        \bottomrule
        \textbf{Method} & \textbf{Latency (ms)}  & \textbf{Speedup} \\
        \midrule
         Dense & 1791  & - \\
        SparseGPT & 1458 & 1.23× \\
        \ourmethod~(ours)&  1450 & 1.24× \\
        \bottomrule
    \end{tabular}}
    \label{tab:speedup}
    \vspace{-2mm}
\end{wraptable}

\textbf{Inference Acceleration.}2:4 sparsity is a common semi-structured pattern, and NVIDIA’s CUTLASS library provides optimized kernels for it. Results of end-to-end latency on LLaMA2-70B can be found in Table~\ref{tab:speedup}. The inference results of SparseGPT and \ourmethod~are nearly identical. \ourmethod~treats every group of four weights as a unit and reorders them when handling 2:4 sparsity, without altering the standard 2:4 sparsity pattern. The reorder operation is conducted during pruning (shown in Appendix~\ref{sec:alg}), and no extra reorder operations are needed during inference after pruning. Consequently, both versions achieve similar acceleration.

\section{Conclusion}
This paper introduces \ourmethod, a new one-shot layerwise pruning method based on the second-order pruning framework. We are motivated by an interesting observation that certain layers in existing LLMs exhibit a \textit{columnar} pattern, and directly applying SparseGPT leads to suboptimal results. 
We propose \ourmethod~to reorder these layers, allowing columns with higher pruning loss to be processed first, thereby preserving more adjustable parameters. 
\ourmethod~employs a pre-pruning step to estimate both column-wise and block-wise pruning loss. It leverages the relative range of block loss to identify \textit{columnar} layers and subsequently performs two-level reordering operations on them. Specifically, within each block, columns are reordered in descending order of individual column loss, while the blocks are reordered in descending order of block loss.
Extensive results on representative LLMs show \ourmethod~surpasses SparseGPT and other counterpart pruning methods.

\bibliography{cpal_2026}

\begin{thebibliography}{51}
\providecommand{\natexlab}[1]{#1}
\providecommand{\url}[1]{\texttt{#1}}
\expandafter\ifx\csname urlstyle\endcsname\relax
  \providecommand{\doi}[1]{doi: #1}\else
  \providecommand{\doi}{doi: \begingroup \urlstyle{rm}\Url}\fi

\bibitem[Frantar and Alistarh(2023)]{sparsegpt}
Elias Frantar and Dan Alistarh.
\newblock Sparsegpt: Massive language models can be accurately pruned in one-shot.
\newblock In \emph{ICML}, 2023.

\bibitem[Thoppilan et~al.(2022)Thoppilan, De~Freitas, Hall, Shazeer, Kulshreshtha, Cheng, Jin, Bos, Baker, Du, et~al.]{thoppilan2022lamda}
Romal Thoppilan, Daniel De~Freitas, Jamie Hall, Noam Shazeer, Apoorv Kulshreshtha, Heng-Tze Cheng, Alicia Jin, Taylor Bos, Leslie Baker, Yu~Du, et~al.
\newblock Lamda: Language models for dialog applications.
\newblock \emph{arXiv preprint arXiv:2201.08239}, 2022.

\bibitem[Achiam et~al.(2023)Achiam, Adler, Agarwal, Ahmad, Akkaya, Aleman, Almeida, Altenschmidt, Altman, Anadkat, et~al.]{achiam2023gpt}
Josh Achiam, Steven Adler, Sandhini Agarwal, Lama Ahmad, Ilge Akkaya, Florencia~Leoni Aleman, Diogo Almeida, Janko Altenschmidt, Sam Altman, Shyamal Anadkat, et~al.
\newblock Gpt-4 technical report.
\newblock \emph{arXiv preprint arXiv:2303.08774}, 2023.

\bibitem[Guo et~al.(2025)Guo, Yang, Zhang, Song, Zhang, Xu, Zhu, Ma, Wang, Bi, et~al.]{deepseek}
Daya Guo, Dejian Yang, Haowei Zhang, Junxiao Song, Ruoyu Zhang, Runxin Xu, Qihao Zhu, Shirong Ma, Peiyi Wang, Xiao Bi, et~al.
\newblock Deepseek-r1: Incentivizing reasoning capability in llms via reinforcement learning.
\newblock \emph{arXiv preprint arXiv:2501.12948}, 2025.

\bibitem[Yang et~al.(2025)Yang, Li, Yang, Zhang, Hui, Zheng, Yu, Gao, Huang, Lv, et~al.]{qwen3}
An~Yang, Anfeng Li, Baosong Yang, Beichen Zhang, Binyuan Hui, Bo~Zheng, Bowen Yu, Chang Gao, Chengen Huang, Chenxu Lv, et~al.
\newblock Qwen3 technical report.
\newblock \emph{arXiv preprint arXiv:2505.09388}, 2025.

\bibitem[Zhang et~al.(2025)Zhang, Liu, et~al.]{translation}
Yu~Zhang, Chen Liu, et~al.
\newblock Native sparse attention: Accelerating {MT} without sacrificing quality.
\newblock In \emph{ACL}, 2025.

\bibitem[Wang et~al.(2025)Wang, Li, et~al.]{text-summarization}
Li~Wang, Jun Li, et~al.
\newblock {MACP}: Parameter-efficient fine-tuning for abstractive summarization.
\newblock In \emph{ICML}, 2025.

\bibitem[Brown et~al.(2025)Brown, Singh, et~al.]{question-answering}
Alex Brown, Amrita Singh, et~al.
\newblock {EPO}: Multi-round rl pushes an 8{B} model beyond {GPT-4} on open-domain {QA}.
\newblock In \emph{NeurIPS}, 2025.

\bibitem[Feng et~al.(2025)Feng, Fang, Ma, and Wang]{feng2025efficient}
Sicheng Feng, Gongfan Fang, Xinyin Ma, and Xinchao Wang.
\newblock Efficient reasoning models: A survey.
\newblock \emph{arXiv preprint arXiv:2504.10903}, 2025.

\bibitem[Steiner et~al.(2023)Steiner, Elhoushi, Kahn, and Hegarty]{llm-challenge}
Benoit Steiner, Mostafa Elhoushi, Jacob Kahn, and James Hegarty.
\newblock Model: memory optimizations for deep learning.
\newblock In \emph{ICML}, 2023.

\bibitem[Luo et~al.(2018)Luo, Zhang, Zhou, Xie, Wu, and Lin]{retrain-1}
Jian-Hao Luo, Hao Zhang, Hong-Yu Zhou, Chen-Wei Xie, Jianxin Wu, and Weiyao Lin.
\newblock Thinet: Pruning cnn filters for a thinner net.
\newblock \emph{TPAMI}, 41\penalty0 (10):\penalty0 2525--2538, 2018.

\bibitem[Bai et~al.(2025)Bai, Jian, Liang, Yin, and Wang]{bai2025ressvd}
Haolei Bai, Siyong Jian, Tuo Liang, Yu~Yin, and Huan Wang.
\newblock Ressvd: Residual compensated svd for large language model compression.
\newblock \emph{arXiv preprint arXiv:2505.20112}, 2025.

\bibitem[LeCun et~al.(1990)LeCun, Denker, and Solla]{OBD}
Y.~LeCun, J.~S. Denker, and S.~A. Solla.
\newblock Optimal brain damage.
\newblock In \emph{NeurIPS}, 1990.

\bibitem[Hassibi et~al.(1993)Hassibi, Stork, and Wolff]{OBS}
B.~Hassibi, D.G. Stork, and G.J. Wolff.
\newblock Optimal brain surgeon and general network pruning.
\newblock In \emph{NeurIPS}, 1993.

\bibitem[Han et~al.(2016)Han, Mao, and Dally]{magnitude}
Song Han, Huizi Mao, and William~J Dally.
\newblock Deep compression: Compressing deep neural networks with pruning, trained quantization and huffman coding.
\newblock In \emph{ICLR}, 2016.

\bibitem[Zhu et~al.(2025)Zhu, Wang, Su, Wang, and Wang]{zhu2025obs}
Junhan Zhu, Hesong Wang, Mingluo Su, Zefang Wang, and Huan Wang.
\newblock Obs-diff: Accurate pruning for diffusion models in one-shot.
\newblock \emph{arXiv preprint arXiv:2510.06751}, 2025.

\bibitem[Feng et~al.(2024)Feng, Tao, and Wang]{feng2024oracle}
Sicheng Feng, Keda Tao, and Huan Wang.
\newblock Is oracle pruning the true oracle?
\newblock \emph{arXiv preprint arXiv:2412.00143}, 2024.

\bibitem[Tuo and Wang(2025)]{tuo2025sparsessm}
Kaiwen Tuo and Huan Wang.
\newblock Sparsessm: Efficient selective structured state space models can be pruned in one-shot.
\newblock \emph{arXiv preprint arXiv:2506.09613}, 2025.

\bibitem[Molchanov et~al.(2019)Molchanov, Mallya, Tyree, Frosio, and Kautz]{global-talor}
Pavlo Molchanov, Arun Mallya, Stephen Tyree, Iuri Frosio, and Jan Kautz.
\newblock Importance estimation for neural network pruning.
\newblock In \emph{CVPR}, 2019.

\bibitem[Yang et~al.(2023)Yang, Wang, Zeng, Li, and Gao]{global-hessian}
Yue Yang, Zi~Wang, Xiaozhe Zeng, Zhengxue Li, and Xinbo Gao.
\newblock Global vision transformer pruning with hessian-aware saliency.
\newblock In \emph{CVPR}, 2023.

\bibitem[Li et~al.(2017)Li, Kadav, Durdanovic, Samet, and Graf]{global-fiter-retraining}
Hao Li, Asim Kadav, Igor Durdanovic, Hanan Samet, and Hans~Peter Graf.
\newblock Pruning filters for efficient convnets.
\newblock In \emph{ICLR}, 2017.

\bibitem[Liu et~al.(2017)Liu, Li, Shen, Huang, Yan, and Zhang]{itertive-channel}
Zhuang Liu, Jianguo Li, Zhiqi Shen, Gao Huang, Shoumeng Yan, and Changshui Zhang.
\newblock Learning efficient convolutional networks through network slimming.
\newblock In \emph{ICCV}, 2017.

\bibitem[Han et~al.(2015)Han, Pool, Tran, and Dally]{magnitude-itertive}
Song Han, Jeff Pool, John Tran, and William Dally.
\newblock Learning both weights and connections for efficient neural network.
\newblock In \emph{NeurIPS}, 2015.

\bibitem[Zhang et~al.(2023)Zhang, Wang, Liu, and Liu]{itertive-magnitude-retrain}
Y.~Zhang, X.~Wang, Y.~Liu, and J.~Liu.
\newblock Magnitude attention-based dynamic pruning.
\newblock In \emph{ICML}, 2023.

\bibitem[Hoefler et~al.(2022)Hoefler, Alistarh, Ben-Nun, Dryden, and Peste]{itertive-Growth}
Torsten Hoefler, Dan Alistarh, Tal Ben-Nun, Nikoli Dryden, and Alexandra Peste.
\newblock Sparsity in deep learning: Pruning and growth for efficient inference and training.
\newblock \emph{JMLR}, 22\penalty0 (241):\penalty0 1--124, 2022.

\bibitem[Ling et~al.(2024)Ling, Wang, and Liu]{slimgpt}
Gui Ling, Ziyang Wang, and Qingwen Liu.
\newblock Slimgpt: Layer-wise structured pruning for large language models.
\newblock In \emph{NeurIPS}, 2024.

\bibitem[Sun et~al.(2023)Sun, Liu, Bair, and Kolter]{wanda}
Mingjie Sun, Zhuang Liu, Anna Bair, and J~Zico Kolter.
\newblock A simple and effective pruning approach for large language models.
\newblock \emph{arXiv preprint arXiv:2306.11695}, 2023.

\bibitem[Zhang et~al.(2024)Zhang, Zhao, Lin, Sun, Yao, Han, Tanner, Liu, and Ji]{dsnot}
Yuxin Zhang, Lirui Zhao, Mingbao Lin, Yunyun Sun, Yiwu Yao, Xingjia Han, Jared Tanner, Shiwei Liu, and Rongrong Ji.
\newblock Dynamic sparse no training: Training-free fine-tuning for sparse llms.
\newblock In \emph{ICLR}, 2024.

\bibitem[Frantar and Alistarh(2022)]{frantar2022obc}
Elias Frantar and Dan Alistarh.
\newblock Optimal brain compression: A framework for accurate post-training quantization and pruning.
\newblock In \emph{NeurIPS}, 2022.

\bibitem[Singh and Alistarh(2020)]{woodfisher}
Sidak~Pal Singh and Dan Alistarh.
\newblock Woodfisher: Efficient second-order approximations for model compression.
\newblock In \emph{NeurIPS}, 2020.

\bibitem[Michel et~al.(2019)Michel, Levy, and Neubig]{head-sixteen}
Paul Michel, Omer Levy, and Graham Neubig.
\newblock Are sixteen heads really better than one?
\newblock In \emph{NeurIPS}, 2019.

\bibitem[Fang et~al.(2023)Fang, Ma, Song, Mi, and Wang]{depgraph}
Gongfan Fang, Xinyin Ma, Mingli Song, Michael~Bi Mi, and Xinchao Wang.
\newblock Depgraph: Towards any structural pruning.
\newblock In \emph{CVPR}, 2023.

\bibitem[Kang and Han(2020)]{channel-pruning}
Minsoo Kang and Bohyung Han.
\newblock Operation-aware soft channel pruning using differentiable masks.
\newblock In \emph{ICML}, 2020.

\bibitem[Zhang and Papyan(2025)]{oats}
Stephen Zhang and Vardan Papyan.
\newblock Oats: Outlier-aware pruning through sparse and low rank decomposition.
\newblock In \emph{ICLR}, 2025.

\bibitem[Dettmers et~al.(2022)Dettmers, Lewis, Belkada, and Zettlemoyer]{outiler-gpt-int8}
Tim Dettmers, Mike Lewis, Younes Belkada, and Luke Zettlemoyer.
\newblock Gpt3. int8 (): 8-bit matrix multiplication for transformers at scale.
\newblock In \emph{NeurIPS}, 2022.

\bibitem[Shin et~al.(2024)Shin, Park, Lee, and Lee]{shin2024rethinking}
Sungbin Shin, Wonpyo Park, Jaeho Lee, and Namhoon Lee.
\newblock Rethinking pruning large language models: Benefits and pitfalls of reconstruction error minimization.
\newblock In \emph{EMNLP}, 2024.

\bibitem[Frantar et~al.(2022)Frantar, Ashkboos, Hoefler, and Alistarh]{frantar2022gptq}
Elias Frantar, Saleh Ashkboos, Torsten Hoefler, and Dan Alistarh.
\newblock Gptq: Accurate post-training quantization for generative pre-trained transformers.
\newblock In \emph{ICLR}, 2022.

\bibitem[Touvron et~al.(2023)Touvron, Martin, Stone, Albert, Almahairi, Babaei, Bashlykov, Batra, Bhargava, Bhosale, et~al.]{touvron2023bllama}
Hugo Touvron, Louis Martin, Kevin Stone, Peter Albert, Amjad Almahairi, Yasmine Babaei, Nikolay Bashlykov, Soumya Batra, Prajjwal Bhargava, Shruti Bhosale, et~al.
\newblock Llama 2: Open foundation and fine-tuned chat models.
\newblock \emph{arXiv preprint arXiv:2307.09288}, 2023.

\bibitem[Dubey et~al.(2024)Dubey, Jauhri, Pandey, Kadian, Al-Dahle, Letman, Mathur, Schelten, Yang, Fan, et~al.]{llama3}
Abhimanyu Dubey, Abhinav Jauhri, Abhinav Pandey, Abhishek Kadian, Ahmad Al-Dahle, Aiesha Letman, Akhil Mathur, Alan Schelten, Amy Yang, Angela Fan, et~al.
\newblock The llama 3 herd of models.
\newblock \emph{arXiv preprint arXiv:2407.21783}, 2024.

\bibitem[Jiang et~al.(2023)Jiang, Sablayrolles, Mensch, Bamford, Chaplot, Casas, Bressand, Lengyel, Lample, Saulnier, et~al.]{mistral}
Albert~Q. Jiang, Alexandre Sablayrolles, Arthur Mensch, Chris Bamford, Devendra~Singh Chaplot, Diego de~las Casas, Florian Bressand, Gianna Lengyel, Guillaume Lample, Lucile Saulnier, et~al.
\newblock Mistral 7b.
\newblock \emph{arXiv preprint arXiv:2310.06825}, 2023.

\bibitem[Merity et~al.(2016)Merity, Xiong, Bradbury, and Socher]{merity2016pointer}
Stephen Merity, Caiming Xiong, James Bradbury, and Richard Socher.
\newblock Pointer sentinel mixture models.
\newblock \emph{arXiv preprint arXiv:1609.07843}, 2016.

\bibitem[Clark et~al.(2019)Clark, Lee, Chang, Kwiatkowski, Collins, and Toutanova]{clark2019boolq}
Christopher Clark, Kenton Lee, Ming-Wei Chang, Tom Kwiatkowski, Michael Collins, and Kristina Toutanova.
\newblock Boolq: Exploring the surprising difficulty of natural yes/no questions.
\newblock \emph{arXiv preprint arXiv:1905.10044}, 2019.

\bibitem[Sakaguchi et~al.(2021)Sakaguchi, Bras, Bhagavatula, and Choi]{sakaguchi2021winogrande}
Keisuke Sakaguchi, Ronan~Le Bras, Chandra Bhagavatula, and Yejin Choi.
\newblock Winogrande: An adversarial winograd schema challenge at scale.
\newblock \emph{Communications of the ACM}, 2021.

\bibitem[Bisk et~al.(2020)Bisk, Zellers, Gao, Choi, et~al.]{piqa}
Yonatan Bisk, Rowan Zellers, Jianfeng Gao, Yejin Choi, et~al.
\newblock Piqa: Reasoning about physical commonsense in natural language.
\newblock In \emph{AAAI}, 2020.

\bibitem[Mihaylov et~al.(2018)Mihaylov, Clark, Khot, and Sabharwal]{obqa}
Todor Mihaylov, Peter Clark, Tushar Khot, and Ashish Sabharwal.
\newblock Can a suit of armor conduct electricity? a new dataset for open book question answering.
\newblock In \emph{EMNLP}, 2018.

\bibitem[Zellers et~al.(2019)Zellers, Holtzman, Bisk, Farhadi, and Choi]{zellers2019hellaswag}
Rowan Zellers, Ari Holtzman, Yonatan Bisk, Ali Farhadi, and Yejin Choi.
\newblock Hellaswag: Can a machine really finish your sentence?
\newblock \emph{arXiv preprint arXiv:1905.07830}, 2019.

\bibitem[Clark et~al.(2018)Clark, Cowhey, Etzioni, Khot, Sabharwal, Schoenick, and Tafjord]{arc}
Peter Clark, Isaac Cowhey, Oren Etzioni, Tushar Khot, Ashish Sabharwal, Carissa Schoenick, and Oyvind Tafjord.
\newblock Think you have solved question answering? try arc, the ai2 reasoning challenge.
\newblock \emph{arXiv preprint arXiv:1803.05457}, 2018.

\bibitem[Sutawika et~al.(2013)Sutawika, Gao, Schoelkopf, Biderman, Tow, Abbasi, Fattori, Lovering, Phang, Thite, et~al.]{lm-eval}
L~Sutawika, L~Gao, H~Schoelkopf, S~Biderman, J~Tow, B~Abbasi, B~Fattori, C~Lovering, J~Phang, A~Thite, et~al.
\newblock Eleutherai/lm-evaluation-harness: Major refactor, 2013.

\bibitem[Paszke et~al.(2019)Paszke, Gross, Massa, Lerer, Bradbury, Chanan, Killeen, Lin, Gimelshein, Antiga, et~al.]{pytorch}
Adam Paszke, Sam Gross, Francisco Massa, Adam Lerer, James Bradbury, Gregory Chanan, Trevor Killeen, Zeming Lin, Natalia Gimelshein, Luca Antiga, et~al.
\newblock Pytorch: An imperative style, high-performance deep learning library.
\newblock In \emph{NeurIPS}, 2019.

\bibitem[Wolf et~al.(2019)Wolf, Debut, Sanh, Chaumond, Delangue, Moi, Cistac, Rault, Louf, Funtowicz, et~al.]{huggingface}
Thomas Wolf, Lysandre Debut, Victor Sanh, Julien Chaumond, Clement Delangue, Anthony Moi, Pierric Cistac, Tim Rault, R{\'e}mi Louf, Morgan Funtowicz, et~al.
\newblock Huggingface's transformers: State-of-the-art natural language processing.
\newblock \emph{arXiv preprint arXiv:1910.03771}, 2019.

\bibitem[Raffel et~al.(2020)Raffel, Shazeer, Roberts, Lee, Narang, Matena, Zhou, Li, and Liu]{c4}
Colin Raffel, Noam Shazeer, Adam Roberts, Katherine Lee, Sharan Narang, Michael Matena, Yanqi Zhou, Wei Li, and Peter~J Liu.
\newblock Exploring the limits of transfer learning with a unified text-to-text transformer.
\newblock \emph{JMLR}, 21\penalty0 (140):\penalty0 1--67, 2020.

\end{thebibliography}
\clearpage
\appendix
This supplementary provides additional details to support the main paper. It is organized as follows:
\begin{itemize}
    \item Section \ref{sec:proof} provides the proof for the formula of fast matrix inversion using Cholesky decomposition in SparseGPT.
    \item Section \ref{sec:alg} illustrates the complete algorithm of \ourmethod.
    \item Section \ref{sec:results} contains additional experimental results, including experiments on the combination of quantization and varying sparsity rates.
    \item Section \ref{sec:Weigh Visualization} analyzes different types of layers in large models, including visualizations and distinctions between \textit{columnar} layers and \textit{non-columnar} layers, and presents the rationale for how \ourmethod~identifies \textit{columnar} layers.
\end{itemize}

\section{Proof of Equation~\ref{eq:cholesky_h-1}}\label{sec:proof}
Equation~\ref{eq:cholesky_h-1} employs Cholesky decomposition to pre-compute and store the inverse Hessian information required for pruning from the first column to the last column of the matrix, thereby eliminating the need for iterative inversion of the Hessian. The formal derivation is provided below.

First, we perform the Cholesky decomposition on $\mathbf{H}^{-1}$. It can be written as:
\begin{equation}
\mathbf{H}^{-1}=\mathbf{LL}^{\top},
\end{equation}
where $\mathbf{L}$ is a lower triangular matrix.
Then, the original Hessian matrix can be denoted as:
\begin{equation}
\mathbf{H}=\left[\mathbf{H}^{-1} \right]^{-1}=\left[\mathbf{L}\mathbf{L}^{\top} \right]^{-1}=\left[\mathbf{L}^{\top} \right]^{-1}\mathbf{L}^{-1} .
\end{equation}
According to matrix block multiplication, the sub-matrix $\mathbf{H}$ starting from the $i$-th row and $i$-th column can be expressed as:
\begin{equation}
    \mathbf{H}_{i:,i:} = \left[\mathbf{L^{\top}}\right]^{-1}_{i:,i:} \mathbf{L}^{-1}_{i:,i:}.
\end{equation}
Based on this equation, we have:
\begin{equation}
\begin{split}
    \left[\mathbf{H}_{i:,i:}\right]^{-1} 
    &= \left[\left[\mathbf{L}^{\top}\right]^{-1}_{i:,i:} \mathbf{L}^{-1}_{i:,i:}\right]^{-1} \\
    &= \left[\mathbf{L}^{-1}_{i:,i:}\right]^{-1} \left[\left[\mathbf{L}^{\top}\right]^{-1}_{i:,i:}\right]^{-1}_.
\end{split}
\label{eq:hii-1}
\end{equation}
We can write the $\mathbf{L}$ and $\mathbf{L}^{-1}$ as block matrices: 
\begin{equation}
\underbrace{
    \begin{bmatrix}
        \mathbf{L}_{:i,:i} & O \\
        \mathbf{L}^{*}     & \mathbf{L}_{i:,i:}
    \end{bmatrix}
}_{\mathbf{L}}
\cdot
\underbrace{
    \begin{bmatrix}
        \mathbf{L}_{:i,:i}^{-1} & O \\
        \mathbf{L}^{*}          & \mathbf{L}_{i:,i:}^{-1}
    \end{bmatrix}
}_{\mathbf{L}^{-1}} = \mathbf{I}_n. 
\end{equation}
where $\mathbf{I}_n$ is unit matrix with dimension $n$.
The inverse matrix on the diagonal of a block diagonal matrix can be written as:
\begin{equation}
\left[\mathbf{L}_{i:,i:}^{-1}\right]^{-1}=\mathbf{L}_{i:,i:}.
    \label{eq:l}
\end{equation}

Similarly, for the upper triangular matrix $\mathbf{L}^\top$, we can obtain:
\begin{equation}
\left[\left[\mathbf{L}^{\top}\right]^{-1}_{i:,i:}\right]^{-1}=\mathbf{L}^\top_{i:,i:}.
    \label{eq:lt}
\end{equation}
Substituting Equation~\ref{eq:l} and Equation~\ref{eq:lt} into Equation~\ref{eq:hii-1} yields Equation~\ref{eq:cholesky_h-1}.

Overall, Equation~\ref{eq:cholesky_h-1} pre-stores all Hessian inverse update information during the pruning process, which means the pruning order must be fully determined before pruning is executed. This is why \ourmethod pre-determines the entire pruning order in advance rather than determining it progressively during the pruning process.






\clearpage
\section{\ourmethod~Algorithm}\label{sec:alg}
We present the full procedure of \ourmethod~in Algorithm~\ref{alg:ourmethod}. First, the weight matrix is partitioned into blocks. Within each block, we compute weight importance scores and perform minimum-element selection based on a target sparsity to generate the pruning loss matrix. This loss matrix is then divided into blocks, and the sum of elements within each block is computed as the block-wise loss. Next, the relative range of block loss is calculated. If the value is below a given threshold, we directly apply SparseGPT pruning. Otherwise, the two-step reordering operation is performed: first, columns are reordered within each block according to their loss values; then, all blocks are globally reordered based on their total loss. The activation matrix is correspondingly transformed, and all indices are stored. Subsequently, SparseGPT is applied to prune the reordered weight matrix. Finally, the resulting sparse matrix is restored to its original order using the previously saved indices.
\vspace{5mm}

\begin{algorithm}
\caption{\ourmethod~}
\label{alg:ourmethod}
\begin{algorithmic}[1]
\State \textbf{Input:} Weight matrix $\mathbf{W}$, input activation $\mathbf{X}$, block size $B_s$, target sparsity $p\%$, identification threshold $\eta$
\State \textbf{Output:} Sparse matrix $\mathbf{W}$
\State $K \gets \lceil N / B_s \rceil$
\For{$k = 1$ to $K$}
    \State $i_1 \gets (k-1) \cdot B_s$, $i_2 \gets \min(k \cdot B_s, N)$
    \State $\mathbf{W}^{(k)} \gets \mathbf{W}[:, i_1:i_2]$
    \State $\mathbf{X}^{(k)} \gets \mathbf{X}[:, i_1:i_2]$
    \State $\mathbf{S}_{ij}^{(k)} \gets |\mathbf{W}^{(k)}_{ij}| \cdot \|\mathbf{X}^{(k)}_j\|_2$ 
    \State $\mathbf{L}^{(k)} \gets \textit{p}\% ~\text{smallest scores in}~\mathbf{S}_{ij}^{(k)}$
    \State $L^{(k)} \gets \text{sum}\left(\mathbf{L}^{(k)}\right)$
\EndFor

\State $\text{Calculate relative range} ~\tau ~\text{of}~L;$
\Comment{Equation~\ref{eq:rr}}

\If{$\tau > \eta$}  
    \For{$k = 1$ to $K$}
        \State $i_1 \gets (k-1) \cdot B_s$, $i_2 \gets \min(k \cdot B_s, N)$
        \State $\mathbf{W}^{(k)} \gets \mathbf{W}[:, i_1:i_2]$
        \State $\mathbf{X}^{(k)} \gets \mathbf{X}[:, i_1:i_2]$
        \State $\mathbf{W}^{(k)} \gets \text{Reorder column} \left(\mathbf{W}^{(k)}\right)$
        \Comment{Equation~\ref{eq:reorder-column}}
         \State $\mathbf{X}^{(k)} \gets \text{Reorder column} \left(\mathbf{X}^{(k)}\right)$
    \EndFor
\State $\mathbf{W} \gets \text{Reorder block}\left(\mathbf{W}\right)$
\Comment{Equation~\ref{eq:reorder-block}}
\State $\mathbf{X} \gets \text{Reorder block}\left(\mathbf{X}\right)$

\State $\mathbf{W} \gets \text{SparseGPT}\left(\mathbf{W},\mathbf{X},B_S,p\%\right)$
\State $\mathbf{W} \gets \text{Reorder back}\left(\mathbf{W}\right)$
\Else
\State $\mathbf{W} \gets \text{SparseGPT}\left(\mathbf{W},\mathbf{X},B_S,p\%\right)$
\EndIf
\State \Return $\mathbf{W}$
\end{algorithmic}
\end{algorithm}

\clearpage
\section{More Experimental Results}\label{sec:results}
\subsection{Combining Quantization}

We have conducted additional experiments on sparsification-quantization joint compression. Specifically, we modify the block loss to be the sum of the pre-pruning loss and the pre-quantization loss, while keeping all other components of our method unchanged. The results of the joint compression under 4-bit and 8-bit quantization across varying sparsity levels are shown in Table~\ref{tab:joint-4bit} and Table~\ref{tab:joint-8bit}.
It can be observed that when combined with 4-bit quantization, our method achieves obvious improvements over SparseGPT in zero-shot accuracy across different models and sparsity levels. Under 8-bit quantization, our method outperforms SparseGPT in the vast majority of cases.
This demonstrates the correctness of the core idea of our method and its potential for extension to other compression domains.

\begin{table}[htbp]
\centering
\scriptsize
\caption{Performance of zero-shot task accuracy (\textcolor{red}{$\uparrow$}) on different models under \textbf{4-bit} quantization at different sparsity rates.}
\vspace{2mm}
\label{tab:joint-4bit}
\setlength{\tabcolsep}{3pt}
\resizebox{\textwidth}{!}{
\begin{tabular}{c c c cccccccc c}
\toprule
\textbf{Group} & \textbf{Sparsity} & \textbf{Method} & \textbf{BoolQ} & \textbf{WinoG.} & \textbf{PIQA} & \textbf{OBQA} & \textbf{HellaS.} & \textbf{ARC-e} & \textbf{ARC-c} & \textbf{Avg.} \\
\midrule
\multirow{6}{*}{LLaMA2-7B} 

& \multirow{2}{*}{0\%} & SparseGPT & \textbf{76.42} & \textbf{68.19} & \textbf{78.51} & 42.20 & \textbf{74.60} & 71.97 & 44.28 & \textbf{65.14} \\
& & \cellcolor{mycolor}\ourmethod~(ours) & \cellcolor{mycolor}75.84 & \cellcolor{mycolor}67.88 & \cellcolor{mycolor}78.45 & \cellcolor{mycolor}\textbf{42.40} & \cellcolor{mycolor}\textbf{74.60} & \cellcolor{mycolor}\textbf{72.22} & \cellcolor{mycolor}\textbf{44.62} & \cellcolor{mycolor}\textbf{65.14}  \\
\cmidrule{2-11}
& \multirow{2}{*}{25\%} & SparseGPT & \textbf{74.92} & \textbf{68.90} & 77.42 & 40.60 & 75.15 & 69.36 & 42.66 & 64.14 \\
& & \cellcolor{mycolor}\ourmethod~(ours) & \cellcolor{mycolor}73.73 & \cellcolor{mycolor}68.51 & \cellcolor{mycolor}\textbf{78.18} & \cellcolor{mycolor}\textbf{40.60} & \cellcolor{mycolor}\textbf{75.38} & \cellcolor{mycolor}\textbf{70.83} & \cellcolor{mycolor}\textbf{46.50} & \cellcolor{mycolor}\textbf{64.81} \\
\cmidrule{2-11}
& \multirow{2}{*}{50\%} & SparseGPT & 73.61 & \textbf{69.06} & 75.90 & \textbf{40.60} & 69.31 & 63.51 & 39.93 & 61.70 \\
& & \cellcolor{mycolor}\ourmethod~(ours) & \cellcolor{mycolor}\textbf{75.99} & \cellcolor{mycolor}68.35 & \cellcolor{mycolor}\textbf{77.48} & \cellcolor{mycolor}\textbf{40.60} & \cellcolor{mycolor}\textbf{69.78} & \cellcolor{mycolor}\textbf{65.45} & \cellcolor{mycolor}\textbf{41.04} & \cellcolor{mycolor}\textbf{62.67} \\
\midrule
\multirow{6}{*}{Mistral-7B} 

& \multirow{2}{*}{0\%} & SparseGPT & 80.61 & 71.43 & 80.30 & 43.20 & 77.43 & 74.66 & 50.00 & 68.23 \\
& & \cellcolor{mycolor}\ourmethod~(ours) & \cellcolor{mycolor}\textbf{80.89} & \cellcolor{mycolor}\textbf{72.77} & \cellcolor{mycolor}\textbf{81.34} & \cellcolor{mycolor}\textbf{43.20} & \cellcolor{mycolor}\textbf{78.78} & \cellcolor{mycolor}\textbf{76.01} & \cellcolor{mycolor}\textbf{50.17} & \cellcolor{mycolor}\textbf{69.02} \\
\cmidrule{2-11}
& \multirow{2}{*}{25\%} & SparseGPT & \textbf{81.44} & 69.93 & 79.71 & 43.20 & 77.35 & 73.61 & 47.35 & 67.51 \\
& & \cellcolor{mycolor}\ourmethod~(ours) & \cellcolor{mycolor}80.64 & \cellcolor{mycolor}\textbf{71.11} & \cellcolor{mycolor}\textbf{80.96} & \cellcolor{mycolor}\textbf{44.00} & \cellcolor{mycolor}\textbf{78.51} & \cellcolor{mycolor}\textbf{75.80} & \cellcolor{mycolor}\textbf{48.98} & \cellcolor{mycolor}\textbf{68.57} \\
\cmidrule{2-11}
& \multirow{2}{*}{50\%} & SparseGPT & 80.86 & 69.38 & 78.13 & \textbf{39.60} & 71.23 & 71.30 & 43.26 & 64.82  \\
& & \cellcolor{mycolor}\ourmethod~(ours) & \cellcolor{mycolor}\textbf{81.96} & \cellcolor{mycolor}\textbf{69.93} & \cellcolor{mycolor}\textbf{78.18} & \cellcolor{mycolor}39.00 & \cellcolor{mycolor}\textbf{72.88} & \cellcolor{mycolor}\textbf{71.93} & \cellcolor{mycolor}\textbf{43.94} & \cellcolor{mycolor}\textbf{65.40} \\

\bottomrule
\end{tabular}
}
\end{table}

\begin{table}[htbp]
\centering
\scriptsize
\caption{Performance of zero-shot task accuracy (\textcolor{red}{$\uparrow$}) on different models under \textbf{8-bit} quantization at different sparsity rates.}
\vspace{2mm}
\label{tab:joint-8bit}
\setlength{\tabcolsep}{3pt}
\resizebox{\textwidth}{!}{
\begin{tabular}{c c c cccccccc c}
\toprule
\textbf{Group} & \textbf{Sparsity} & \textbf{Method} & \textbf{BoolQ} & \textbf{WinoG.} & \textbf{PIQA} & \textbf{OBQA} & \textbf{HellaS.} & \textbf{ARC-e} & \textbf{ARC-c} & \textbf{Avg.} \\

\midrule
\multirow{6}{*}{LLaMA2-7B} 

& \multirow{2}{*}{0\%} & SparseGPT & 77.34 & \textbf{68.82} & 79.00 & \textbf{44.20} & 76.00 & 74.33 & \textbf{46.25} & 66.56 \\
& & \cellcolor{mycolor}\ourmethod~(ours) & \cellcolor{mycolor}\textbf{77.40} & \cellcolor{mycolor}68.75 & \cellcolor{mycolor}\textbf{79.05} & \cellcolor{mycolor}\textbf{44.20} & \cellcolor{mycolor}\textbf{76.01} & \cellcolor{mycolor}\textbf{74.54} & \cellcolor{mycolor}46.16 & \cellcolor{mycolor}\textbf{66.59}  \\
\cmidrule{2-11}
& \multirow{2}{*}{25\%} & SparseGPT & 76.97 & 69.61 & 78.84 & 44.60 &\textbf{76.26} & \textbf{73.06} & \textbf{46.50} & 66.55 \\
& & \cellcolor{mycolor}\ourmethod~(ours) & \cellcolor{mycolor}\textbf{77.03} & \cellcolor{mycolor}\textbf{70.24} & \cellcolor{mycolor}\textbf{79.16} & \cellcolor{mycolor}\textbf{45.00} & \cellcolor{mycolor}76.25 & \cellcolor{mycolor}72.43 & \cellcolor{mycolor}46.33 & \cellcolor{mycolor}\textbf{66.63} \\
\cmidrule{2-11}
& \multirow{2}{*}{50\%} & SparseGPT  & \textbf{75.78} & \textbf{70.17} & \textbf{77.15} & 41.80 & \textbf{71.11} & \textbf{67.85} & 41.47 & \textbf{63.62} \\
& & \cellcolor{mycolor}\ourmethod~(ours)  & \cellcolor{mycolor}75.54 & \cellcolor{mycolor}69.30 & \cellcolor{mycolor}76.88 & \cellcolor{mycolor}\textbf{42.80} & \cellcolor{mycolor}70.85 & \cellcolor{mycolor}66.54 & \cellcolor{mycolor}\textbf{41.55} & \cellcolor{mycolor}63.35 \\
\midrule
\multirow{6}{*}{Mistral-7B} 

& \multirow{2}{*}{0\%} & SparseGPT & \textbf{81.99} & \textbf{73.88} & 81.77 & \textbf{45.00} & \textbf{80.40} & \textbf{78.49} & \textbf{52.30} & \textbf{70.55} \\
& & \cellcolor{mycolor}\ourmethod~(ours)  & \cellcolor{mycolor}81.96 & \cellcolor{mycolor}73.48 & \cellcolor{mycolor}\textbf{81.94} & \cellcolor{mycolor}44.80 & \cellcolor{mycolor}80.36 & \cellcolor{mycolor}78.45 & \cellcolor{mycolor}52.30 & \cellcolor{mycolor}70.47  \\
\cmidrule{2-11}
& \multirow{2}{*}{25\%} & SparseGPT & \textbf{82.63} & 72.77 & \textbf{81.99} & 43.60 & \textbf{80.12} & 77.78 & \textbf{51.96} & 70.12 \\
& & \cellcolor{mycolor}\ourmethod~(ours) & \cellcolor{mycolor}82.51 & \cellcolor{mycolor}\textbf{73.56} & \cellcolor{mycolor}81.94 & \cellcolor{mycolor}\textbf{43.80} & \cellcolor{mycolor}80.06 & \cellcolor{mycolor}\textbf{77.95} & \cellcolor{mycolor}51.79 & \cellcolor{mycolor}\textbf{70.23} \\
\cmidrule{2-11}
& \multirow{2}{*}{50\%} & SparseGPT & 82.17 & 70.64 & \textbf{79.33} & \textbf{40.80} & \textbf{75.29} & \textbf{73.99} & 44.54 & 66.68  \\
& & \cellcolor{mycolor}\ourmethod~(ours) & \cellcolor{mycolor}\textbf{82.45} & \cellcolor{mycolor}\textbf{70.72} & \cellcolor{mycolor}\textbf{79.33} & \cellcolor{mycolor}40.60 & \cellcolor{mycolor}75.25 & \cellcolor{mycolor}73.91 & \cellcolor{mycolor}\textbf{45.82} & \cellcolor{mycolor}\textbf{66.87} \\

\bottomrule
\end{tabular}
}
\end{table}

\clearpage

\subsection{Varying Sparsity Rates}
We extend our evaluation to broader sparsity regimes in Table~\ref{tab:sparsity-llama3-8b} and~\ref{tab:sparsity-mistral-7b}. While both methods yield competitive results under moderate sparsity, our method exhibits increasingly clear advantages over SparseGPT at higher sparsity rates.

\begin{table}[htbp]
\centering
\scriptsize
\caption{Performance of zero-shot task accuracy~(\textcolor{red}{$\uparrow$}) for different unstructured pruning methods on LLaMA3-8B at different sparsity rates. }
\label{tab:sparsity-llama3-8b}
\vspace{2mm}
\setlength{\tabcolsep}{4pt}
\resizebox{0.96\textwidth}{!}{
\begin{tabular}{c cccccccc c}
\toprule
\textbf{Sparsity} & \textbf{Method} & \textbf{BoolQ} & \textbf{WinoG.} & \textbf{PIQA} & \textbf{OBQA} & \textbf{HellaS.} & \textbf{ARC-e} & \textbf{ARC-c} & \textbf{Avg.} \\
\midrule
0\% & Dense & 81.38 & 72.61 & 80.79 & 45.00 & 79.16 & 77.74 & 53.24 & 69.99  \\
\midrule
\multirow{2}{*}{30\%} 
& SparseGPT & 82.14 & 74.03 & 80.09 & \textbf{44.20} & 78.54 & \textbf{76.60} & 50.43 & 69.43 \\
& \cellcolor{mycolor}\ourmethod~(ours) & \cellcolor{mycolor}\textbf{82.23} & \cellcolor{mycolor}\textbf{74.51} & \cellcolor{mycolor}\textbf{80.58} & \cellcolor{mycolor}\textbf{44.20} & \cellcolor{mycolor}\textbf{78.56} & \cellcolor{mycolor}76.43 & \cellcolor{mycolor}\textbf{50.51} & \cellcolor{mycolor}\textbf{69.57}\\
\midrule

\multirow{2}{*}{40\%} 
& SparseGPT & \textbf{82.42} & \textbf{73.24} & \textbf{79.43} & 42.80 & \textbf{76.71} & \textbf{74.87} & 50.09 & 68.51 \\
& \cellcolor{mycolor}\ourmethod~(ours) & \cellcolor{mycolor}82.14 & \cellcolor{mycolor}72.61 & \cellcolor{mycolor}79.00 & \cellcolor{mycolor}\textbf{44.60} & \cellcolor{mycolor}76.62 & \cellcolor{mycolor}74.37 & \cellcolor{mycolor}\textbf{50.34} & \cellcolor{mycolor}\textbf{68.53} \\
\midrule

\multirow{2}{*}{50\%} 
& SparseGPT & 78.41 & \textbf{72.85} & \textbf{77.64} & \textbf{41.60} & 72.92 & 68.73 & 44.11 & 65.18 \\
& \cellcolor{mycolor}\ourmethod~(ours) & \cellcolor{mycolor}\textbf{78.87} & \cellcolor{mycolor}72.45 & \cellcolor{mycolor}77.20 & \cellcolor{mycolor}40.80 & \cellcolor{mycolor}\textbf{73.19} & \cellcolor{mycolor}\textbf{70.92} & \cellcolor{mycolor}\textbf{45.65} & \cellcolor{mycolor}\textbf{65.58} \\
\midrule

\multirow{2}{*}{60\%} 
& SparseGPT & 75.78 & \textbf{68.43} & \textbf{72.52} & \textbf{37.40} & \textbf{61.79} & \textbf{59.72} & \textbf{34.64} & \textbf{58.61} \\
& \cellcolor{mycolor}\ourmethod~(ours) & \cellcolor{mycolor}\textbf{76.39}  & \cellcolor{mycolor}66.69  & \cellcolor{mycolor}72.20  & \cellcolor{mycolor}35.00 &  \cellcolor{mycolor}61.61  & \cellcolor{mycolor}57.58 &  \cellcolor{mycolor}33.19  &\cellcolor{mycolor}57.52 \\
\midrule

\multirow{2}{*}{70\%} 
& SparseGPT & 68.78 & 55.88 & 61.15 & \textbf{29.60} & \textbf{41.12} & 40.11 & \textbf{25.34} & 46.00  \\
& \cellcolor{mycolor}\ourmethod~(ours) & \cellcolor{mycolor}\textbf{68.65} & \cellcolor{mycolor}\textbf{57.14} & \cellcolor{mycolor}\textbf{61.70} & \cellcolor{mycolor}28.80 & \cellcolor{mycolor}40.60 & \cellcolor{mycolor}\textbf{40.74} & \cellcolor{mycolor}24.83 & \cellcolor{mycolor}\textbf{46.07} \\
\midrule

\multirow{2}{*}{80\%} 
& SparseGPT & 53.76 & 49.25 & \textbf{53.48} & 25.60 & \textbf{28.37} & \textbf{30.30} & 20.31 & 37.30 \\
& \cellcolor{mycolor}\ourmethod~(ours) & \cellcolor{mycolor}\textbf{56.67} & \cellcolor{mycolor}\textbf{50.59} & \cellcolor{mycolor}\textbf{53.48} & \cellcolor{mycolor}\textbf{27.00} & \cellcolor{mycolor}28.36 & \cellcolor{mycolor}30.01 & \cellcolor{mycolor}\textbf{21.76} & \cellcolor{mycolor}\textbf{38.27} \\
\midrule

\multirow{2}{*}{90\%} 
& SparseGPT & \textbf{38.07} & 48.62 & 51.52 & 25.00 & \textbf{27.07} & \textbf{28.49} & \textbf{23.29} & 34.58 \\
& \cellcolor{mycolor}\ourmethod~(ours) & \cellcolor{mycolor}37.83 & \cellcolor{mycolor}\textbf{51.30} & \cellcolor{mycolor}\textbf{52.94} & \cellcolor{mycolor}\textbf{27.20} & \cellcolor{mycolor}26.99 & \cellcolor{mycolor}28.07 & \cellcolor{mycolor}23.04 & \cellcolor{mycolor}\textbf{35.34} \\
\bottomrule
\end{tabular}
}
\end{table}
\begin{table}[htbp]
\centering

\scriptsize
\caption{Performance of zero-shot task accuracy~(\textcolor{red}{$\uparrow$}) for different unstructured pruning methods on Mistral-7B at different sparsity rates.}
\label{tab:sparsity-mistral-7b}

\vspace{2mm}
\setlength{\tabcolsep}{4pt}
\resizebox{0.95\textwidth}{!}{
\begin{tabular}{c cccccccc c}
\toprule
\textbf{Sparsity} & \textbf{Method} & \textbf{BoolQ} & \textbf{WinoG.} & \textbf{PIQA} & \textbf{OBQA} & \textbf{HellaS.} & \textbf{ARC-e} & \textbf{ARC-c} & \textbf{Avg.} \\
\midrule
0\% & Dense & 82.14 & 73.80 & 82.26 & 44.20 & 80.42 & 78.20 & 52.30 & 70.47 \\
\midrule
\multirow{2}{*}{30\%} 
& SparseGPT & 82.45 & \textbf{72.69} & \textbf{81.66} & 43.00 & \textbf{79.95} & \textbf{77.86} & \textbf{51.79} & \textbf{69.91} \\
& \cellcolor{mycolor}\ourmethod~(ours) & \cellcolor{mycolor}\textbf{82.54} & \cellcolor{mycolor}72.45 & \cellcolor{mycolor}\textbf{81.66} & \cellcolor{mycolor}\textbf{43.20} & \cellcolor{mycolor}\textbf{79.95} & \cellcolor{mycolor}77.53 & \cellcolor{mycolor}51.54 & \cellcolor{mycolor}69.84 \\
\midrule

\multirow{2}{*}{40\%} 
& SparseGPT & 82.75 & 72.45 & 81.34 & 42.60 & 78.49 & \textbf{76.01} & 48.55 & 68.88 \\
& \cellcolor{mycolor}\ourmethod~(ours) & \cellcolor{mycolor}\textbf{82.81} & \cellcolor{mycolor}\textbf{72.77} & \cellcolor{mycolor}\textbf{81.61} & \cellcolor{mycolor}\textbf{42.60} & \cellcolor{mycolor}\textbf{78.63} & \cellcolor{mycolor}75.88 & \cellcolor{mycolor}\textbf{49.49} & \cellcolor{mycolor}\textbf{69.11} \\
\midrule

\multirow{2}{*}{50\%} 
& SparseGPT & \textbf{83.76} & \textbf{72.30} & \textbf{79.54} & \textbf{40.80} & \textbf{75.61} & 74.37 & 44.97 & \textbf{67.34} \\
& \cellcolor{mycolor}\ourmethod~(ours) & \cellcolor{mycolor}82.39 & \cellcolor{mycolor}71.67 & \cellcolor{mycolor}79.43 & \cellcolor{mycolor}40.60 & \cellcolor{mycolor}75.61 & \cellcolor{mycolor}\textbf{74.49} & \cellcolor{mycolor}\textbf{46.42} & \cellcolor{mycolor}67.23 \\
\midrule

\multirow{2}{*}{60\%} 
& SparseGPT & \textbf{77.83} & \textbf{68.11} & \textbf{76.82} & \textbf{38.80} & 67.25 & \textbf{66.58} & \textbf{39.76} & \textbf{62.16} \\
& \cellcolor{mycolor}\ourmethod~ (ours)& \cellcolor{mycolor}76.27 & \cellcolor{mycolor}67.32 & \cellcolor{mycolor}76.39 & \cellcolor{mycolor}38.20 & \cellcolor{mycolor}\textbf{67.47} & \cellcolor{mycolor}66.37 & \cellcolor{mycolor}37.63 & \cellcolor{mycolor}61.38 \\
\midrule

\multirow{2}{*}{70\%} 
& SparseGPT & \textbf{67.25} & 58.25 & 65.56 & \textbf{31.20} & 46.64 & 46.80 & 27.22 & 48.99 \\
& \cellcolor{mycolor}\ourmethod~(ours) & \cellcolor{mycolor}65.14 & \cellcolor{mycolor}\textbf{60.14} & \cellcolor{mycolor}\textbf{66.10} & \cellcolor{mycolor}30.00 & \cellcolor{mycolor}\textbf{46.94} & \cellcolor{mycolor}\textbf{48.78} & \cellcolor{mycolor}\textbf{27.56} & \cellcolor{mycolor}\textbf{49.24} \\
\midrule

\multirow{2}{*}{80\%} 
& SparseGPT & 53.91 & \textbf{50.75} &\textbf{54.13} & \textbf{26.80} & 29.56 & 29.63 & \textbf{20.90} & 37.95 \\
& \cellcolor{mycolor}\ourmethod~(ours) & \cellcolor{mycolor}\textbf{59.76} & \cellcolor{mycolor}50.20 & \cellcolor{mycolor}53.86 & \cellcolor{mycolor}26.60 & \cellcolor{mycolor}\textbf{29.60} & \cellcolor{mycolor}\textbf{30.05} & \cellcolor{mycolor}20.82 & \cellcolor{mycolor}\textbf{38.70} \\
\midrule

\multirow{2}{*}{90\%} 
& SparseGPT & 37.83 & \textbf{48.93} & \textbf{52.29} & 25.80 & \textbf{27.06} & \textbf{28.37} & \textbf{25.09} & \textbf{35.05} \\
& \cellcolor{mycolor}\ourmethod~(ours) & \cellcolor{mycolor}\textbf{37.86} & \cellcolor{mycolor}47.67 & \cellcolor{mycolor}51.80 & \cellcolor{mycolor}\textbf{26.80} & \cellcolor{mycolor}\textbf{27.06} & \cellcolor{mycolor}\textbf{28.37} & \cellcolor{mycolor}24.74 & \cellcolor{mycolor}34.90 \\
\bottomrule
\end{tabular}
}
\end{table}
\clearpage
\section{Layer Analyses}\label{sec:Weigh Visualization}
\subsection{Columnar Layer Visualization}\label{appendix:Weigh Visualization}
\begin{figure}[!h]
\centering
\setlength{\tabcolsep}{3pt}
\resizebox{\textwidth}{!}{
\begin{tabular}{cccc}
\includegraphics[width=0.25\linewidth]{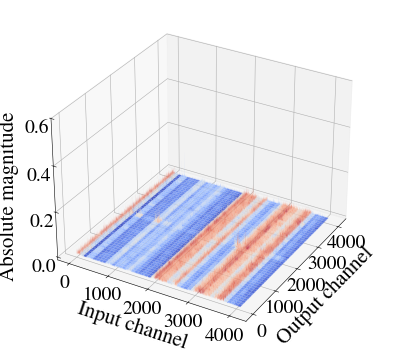} &
\includegraphics[width=0.25\linewidth]{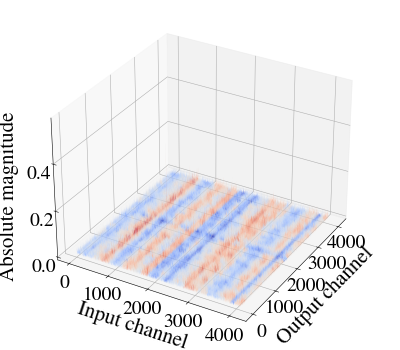} &
\includegraphics[width=0.25\linewidth]{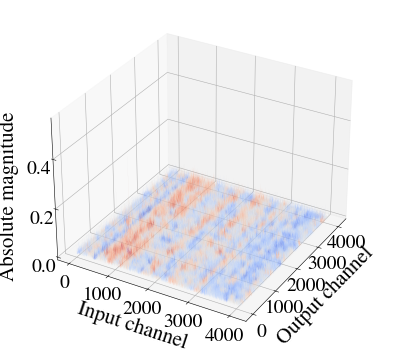} &
\includegraphics[width=0.25\linewidth]{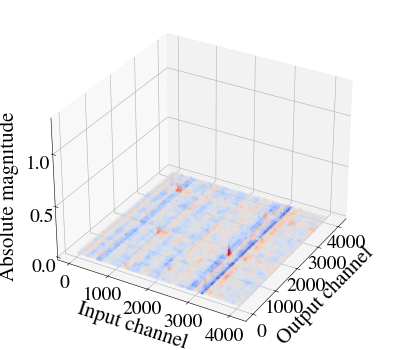} \\
\makecell[l]{\small (a) Layer1 self\_attn.o\_proj} & 
\makecell[l]{\small (b) Layer11 self\_attn.o\_proj} & 
\makecell[l]{\small (c) Layer21 self\_attn.o\_proj} &
\makecell[l]{\small (d) Layer31 self\_attn.o\_proj} 
\\
\end{tabular}
}
\caption{\small Visualization of layers with columnar columnar distribution in \textbf{LLaMA2-7B}}
\label{fig:weight_visualization-llama2-7}
\end{figure}

\begin{figure}[!h]
\centering
\setlength{\tabcolsep}{3pt}
\resizebox{\textwidth}{!}{
\begin{tabular}{cccc}
\includegraphics[width=0.25\linewidth]{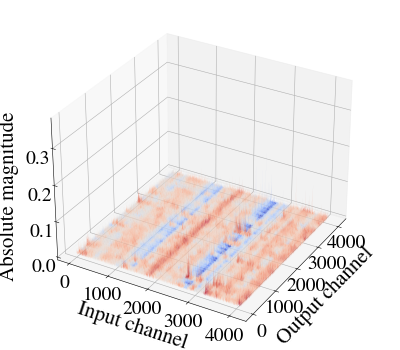} &
\includegraphics[width=0.25\linewidth]{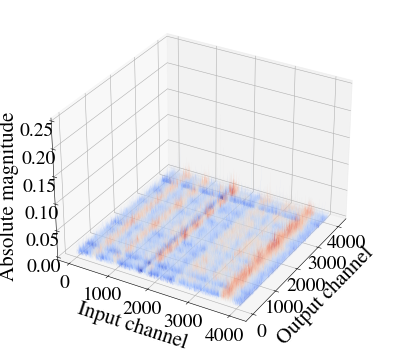} &
\includegraphics[width=0.25\linewidth]{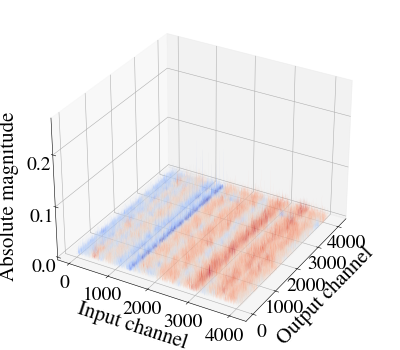} &
\includegraphics[width=0.25\linewidth]{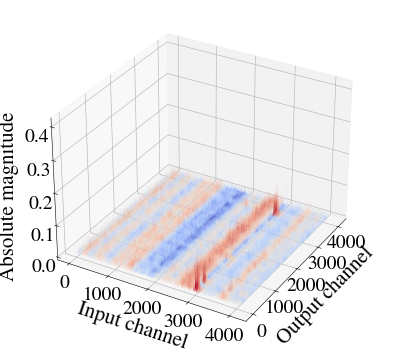} \\
\makecell[l]{\small (a) Layer1 self\_attn.o\_proj} & 
\makecell[l]{\small (b) Layer11 self\_attn.o\_proj} & 
\makecell[l]{\small (c) Layer21 self\_attn.o\_proj} &
\makecell[l]{\small (d) Layer31 self\_attn.o\_proj} \\
\end{tabular}
}
\caption{\small Visualization of layers with the columnar distribution in \textbf{LLaMA3-8B}}
\label{fig:weight_visualization-llama3-8}
\end{figure}

\begin{figure}[!h]
\centering
\setlength{\tabcolsep}{3pt}
\resizebox{\textwidth}{!}{
\begin{tabular}{cccc}
\includegraphics[width=0.25\linewidth]{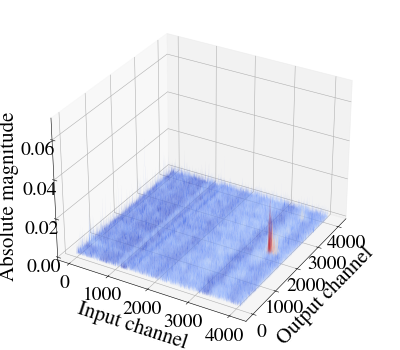} &
\includegraphics[width=0.25\linewidth]{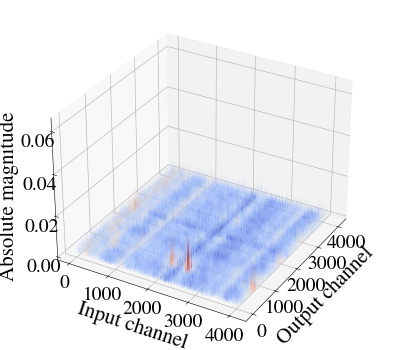} &
\includegraphics[width=0.25\linewidth]{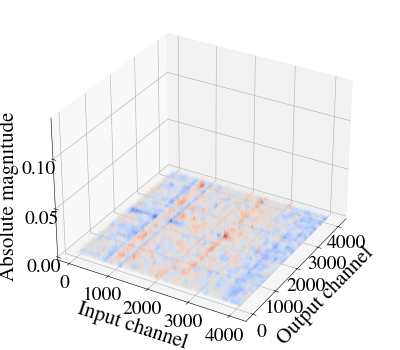} &
\includegraphics[width=0.25\linewidth]{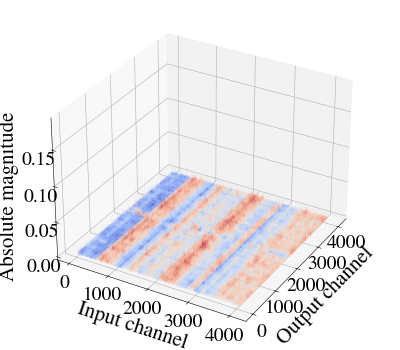} \\
\makecell[l]{\small (a) Layer1 self\_attn.o\_proj} & 
\makecell[l]{\small (b) Layer11 self\_attn.o\_proj} & 
\makecell[l]{\small (c) Layer21 self\_attn.o\_proj} &
\makecell[l]{\small (d) Layer31 self\_attn.o\_proj} \\
\end{tabular}
}

\caption{\small Visualization of layers with a columnar distribution in \textbf{Mistral-7B}}
\label{fig:weight_visualization-mistral}
\end{figure}

Figure~\ref{fig:weight_visualization-llama2-7}-\ref{fig:weight_visualization-mistral} reveal layers with \textit{columnar} pattern in current mainstream LLMs. Specifically, lots of columns with similar magnitudes tend to cluster together.
Moreover, we also find that all matrices exhibiting this pattern are projection matrices of self-attention output.

\begin{figure}[!h]
\centering
\setlength{\tabcolsep}{3pt}
\resizebox{\textwidth}{!}{
\begin{tabular}{cccc}
\includegraphics[width=0.25\linewidth]{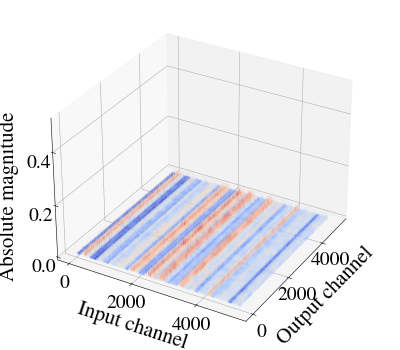} &
\includegraphics[width=0.25\linewidth]{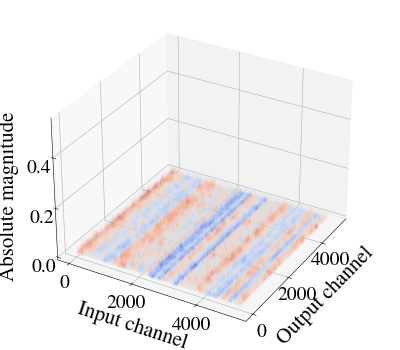} &
\includegraphics[width=0.25\linewidth]{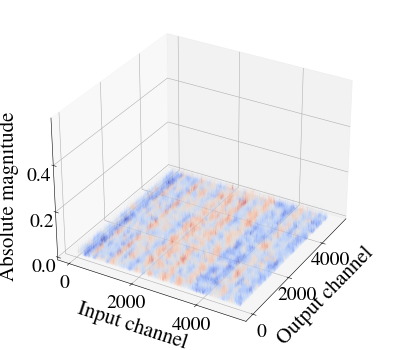} &
\includegraphics[width=0.25\linewidth]{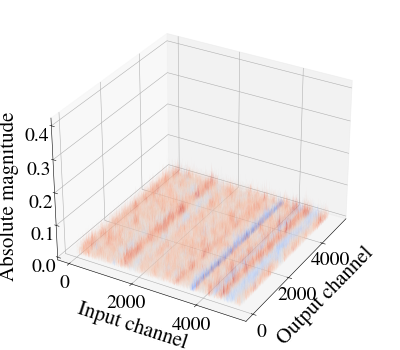} \\
\makecell[l]{\small (a) Layer1 self\_attn.o\_proj} & 
\makecell[l]{\small (b) Layer11 self\_attn.o\_proj} & 
\makecell[l]{\small (c) Layer21 self\_attn.o\_proj} &
\makecell[l]{\small (d) Layer31 self\_attn.o\_proj} \\
\end{tabular}
}
\caption{\small Visualization of layers with tile columnar distribution in \textbf{LLaMA2-13B}}

\end{figure}

\clearpage
\subsection{Non-columar Layer Visualization}\label{appendix:Non-columar Visualization}
\begin{figure}[!h]
     \centering
    \begin{tabular}{ccc}
   \includegraphics[width=0.28\linewidth]{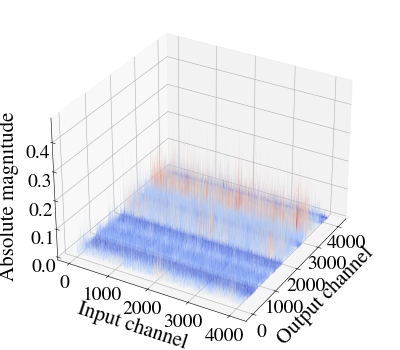} &
    \includegraphics[width=0.28\linewidth]{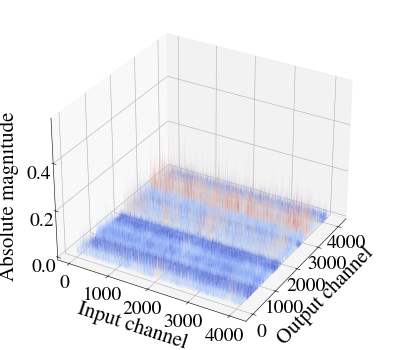} &
    \includegraphics[width=0.28\linewidth]{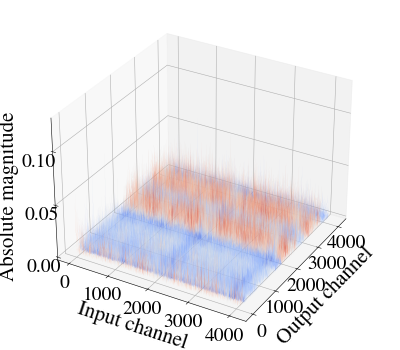} \\
        (a) \small self\_attn.q\_proj 
      &(b) \small self\_attn.k\_proj
      &(c) \small self\_attn.v\_proj \\
      
    \includegraphics[width=0.28\linewidth]{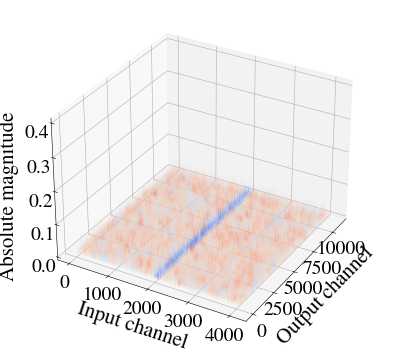} &
    \includegraphics[width=0.28\linewidth]{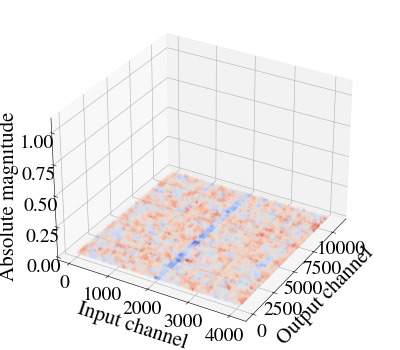} &
    \includegraphics[width=0.28\linewidth]{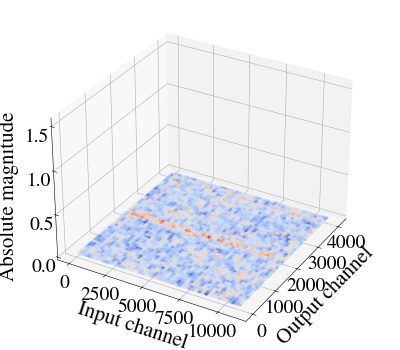} \\
      (d) \small mlp.up\_proj 
      &(e) \small mlp.gate\_proj 
      &(f) \small mlp.down\_proj \\
    \end{tabular}
    
\caption{\small Visualization of non-columar layers in \textbf{LLaMA2-7B}}
\label{fig:llama2-no-columar}
\end{figure}

\begin{figure}[!h]
     \centering
    \begin{tabular}{ccc}
   \includegraphics[width=0.28\linewidth]{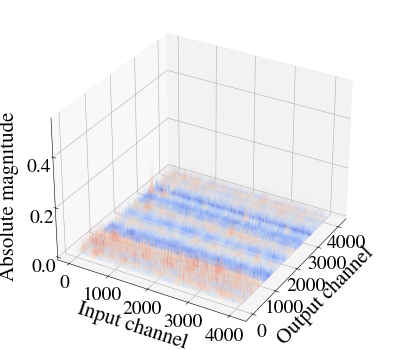} &
    \includegraphics[width=0.28\linewidth]{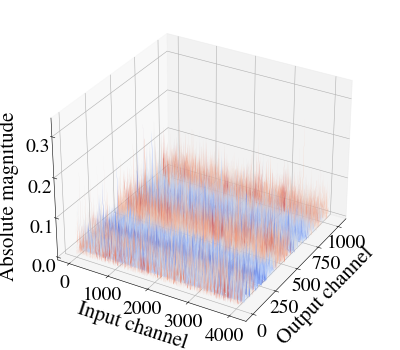} &
    \includegraphics[width=0.28\linewidth]{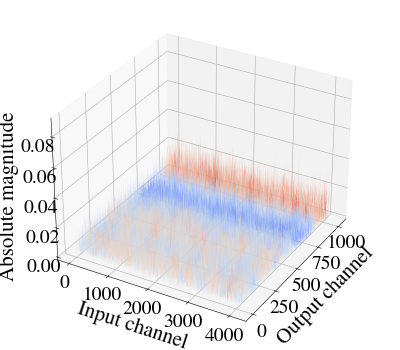} \\
        (a) \small self\_attn.q\_proj 
      &(b) \small self\_attn.k\_proj
      &(c) \small self\_attn.v\_proj \\
      
    \includegraphics[width=0.28\linewidth]{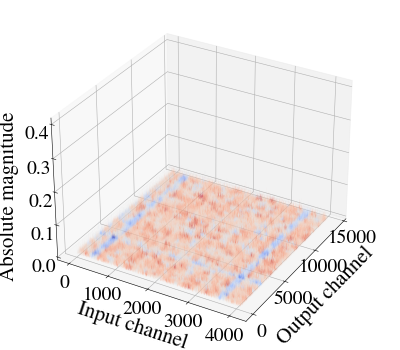} &
    \includegraphics[width=0.28\linewidth]{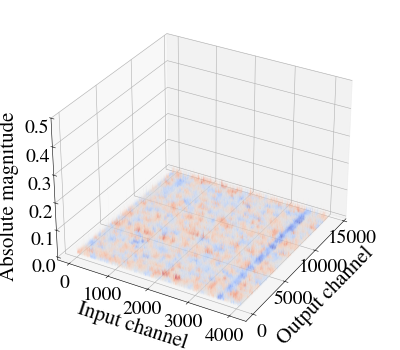} &
    \includegraphics[width=0.28\linewidth]{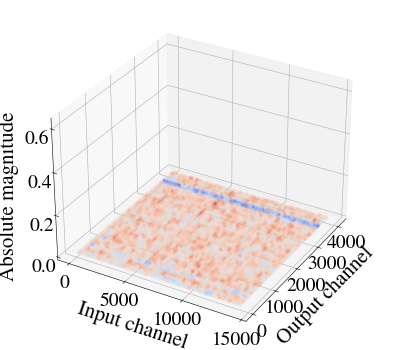} \\
      (d) \small mlp.up\_proj 
      &(e) \small mlp.gate\_proj 
      &(f) \small mlp.down\_proj \\
    \end{tabular}
    
\caption{\small Visualization of non-columar layers in \textbf{LLaMA3-8B}}
\label{fig:llama3-no-columar}
\end{figure}

Figure~\ref{fig:llama2-no-columar} and ~\ref{fig:llama3-no-columar} present a visualization of the \textit{non-columnar} layer weights. It can be observed that the weights of these layers are distributed relatively uniformly along the input channel dimension and no obvious columnar pattern is present.
\vspace{-4mm}

\clearpage
\subsection{Difference between Columar and Non-columnar Layer}
\begin{figure}[!h]
\centering
\resizebox{0.85\textwidth}{!}{
\begin{tabular}{@{}ccc@{}}
     \includegraphics[width=0.4\linewidth]{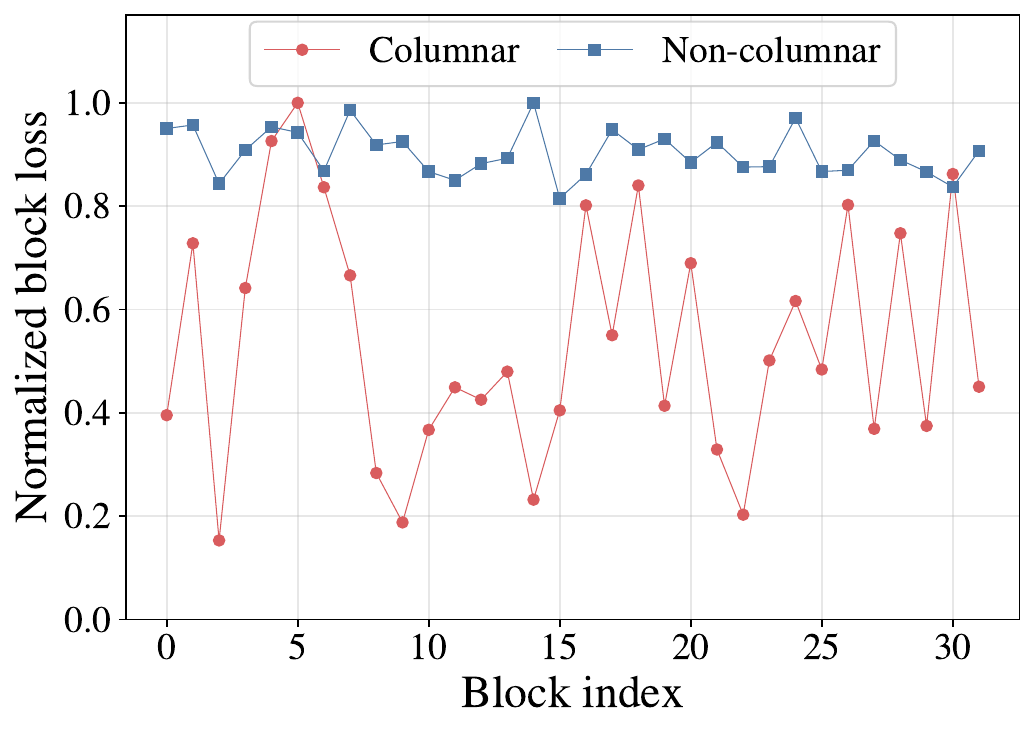} &
     \includegraphics[width=0.4\linewidth]{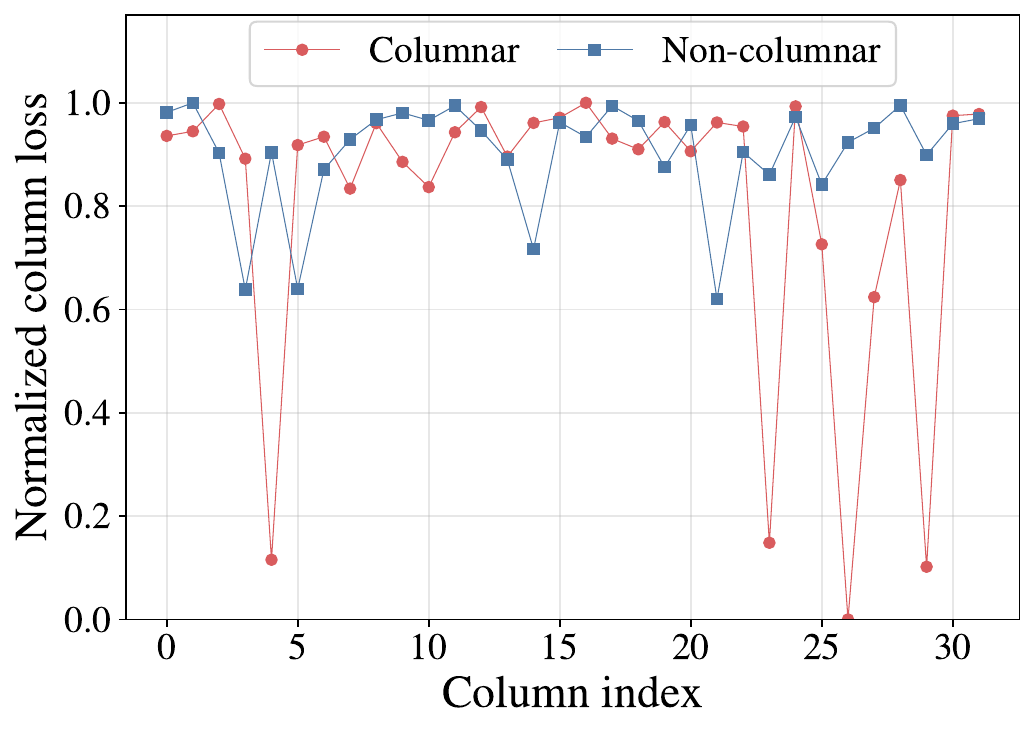} \\ 
     (a) \small Block loss & (b) \small Column loss 
\end{tabular}
}
\caption{\small Difference in block loss and column loss between the columnar and non-columnar layer. For the column loss, we selected 32 consecutive columns in one block for visualization. }
\label{fig:loss}
\end{figure}

 We analyze distributions of column loss and block loss in \textit{columnar} and \textit{non-columnar} layers. Figure~\ref{fig:loss}(a) shows that for \textit{columnar} layers, there is a significant disparity in block loss. For instance, the block with the highest loss is ten times larger than the block with the lowest loss. In contrast, for \textit{non-columnar} layers, due to the relatively uniform weight distribution, the differences in losses between blocks are minimal. Moreover, the variance between different columns within the same block in the columnar layer remains obvious, as shown in Figure~\ref{fig:loss}(b).
 
Overall, for those \textit{columnar} layers, both block loss and column loss exhibit significant fluctuations. This motivates us to adopt a two-stage reordering strategy: block reorder and column reorder, to let weights with potentially greater pruning errors be pruned earlier.


\subsection{Hyperparameter for Identifying Columnar Layer}
\begin{figure}[!h]
\centering
\resizebox{0.85\textwidth}{!}{
\begin{tabular}{@{}cc@{}}
     \includegraphics[width=0.4\linewidth]{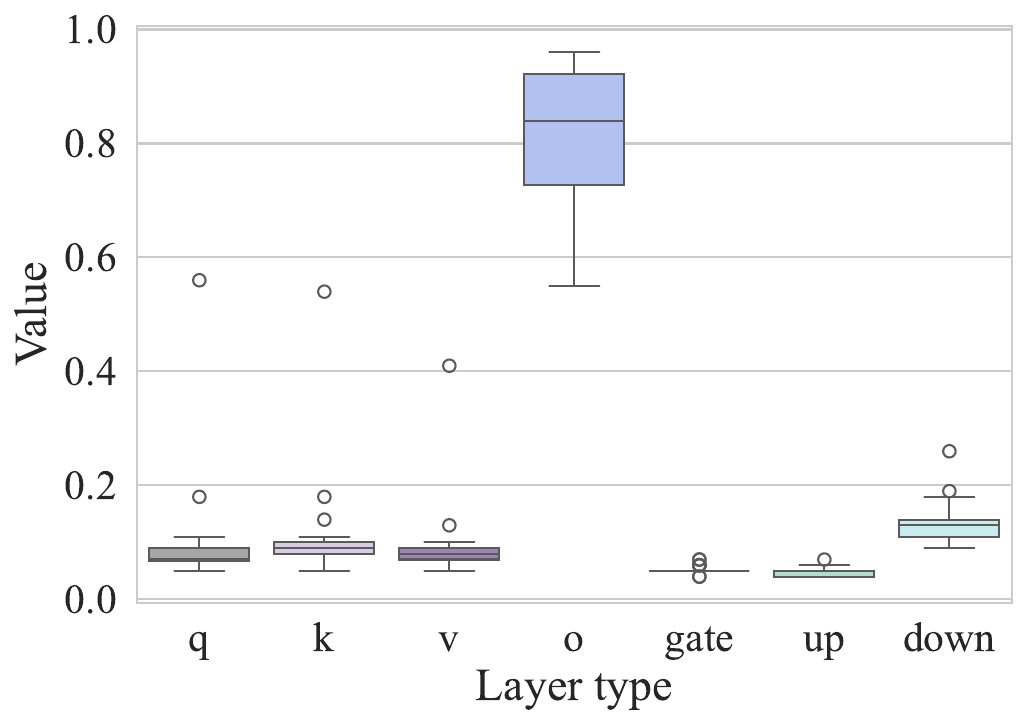} &
     \includegraphics[width=0.4\linewidth]{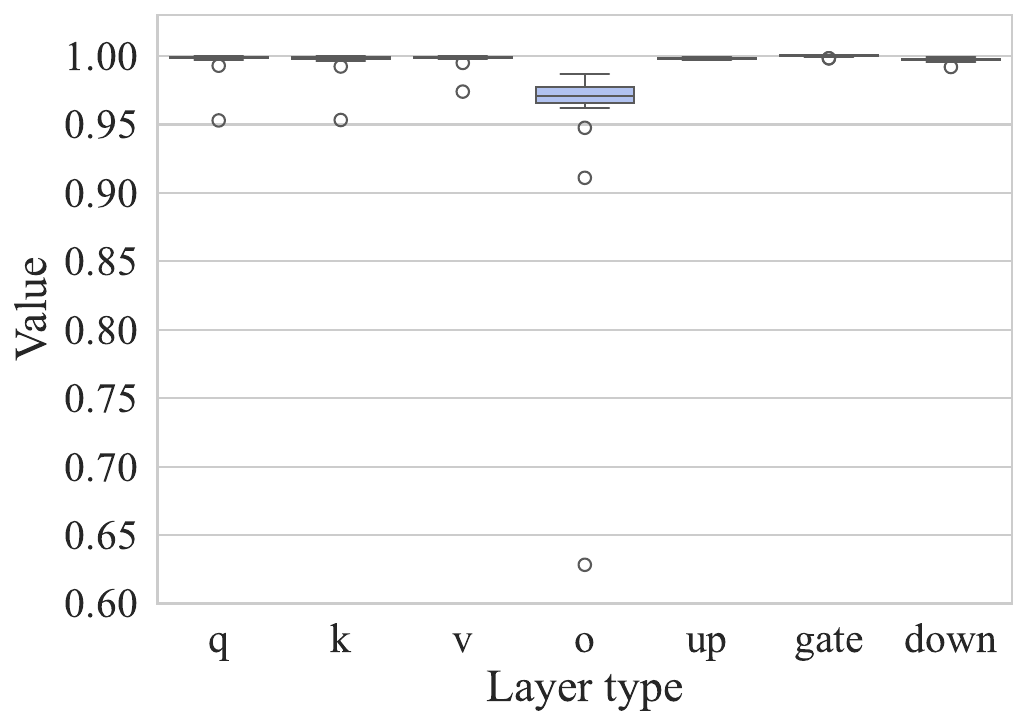}
     \\
     \small (a) Relative range of block loss & \small (b) Relative reconstruction error
\end{tabular}
}
\caption{\small Relative range of block loss and the relative reconstruction error after reordering to that before reordering across layers in LLaMA3-8B at 70\% sparsity.
}
\label{fig:relative-llama3-8b-0.7}
\end{figure}

We conduct a statistical analysis of the relative range of block loss across different types of layers in the whole model. Additionally, we compare the reconstruction error after reordering weight blocks with that of SparseGPT. The statistical results in LLaMA3-8B at 70\% sparsity are shown in Figure~\ref{fig:relative-llama3-8b-0.7}. For all o\_proj layers, the relative range of block loss exceeds 0.5. After applying the reordering strategy, the reconstruction error consistently is reduced. For other layer types, the relative block loss is generally much lower, mostly around 0.1, suggesting a more uniform block loss distribution. After reordering, their reconstruction errors either remain unchanged or exhibit slight reductions.

Based on the above observations, we set the threshold for identifying \textit{columnar} layers to 0.5, ensuring the reconstruction error for identified layers after reordering is consistently reduced.
\vspace{-4mm}

\end{document}